\pdfoutput=1
\documentclass[11pt]{article}

\usepackage[preprint]{acl}

\usepackage{times}
\usepackage{latexsym}
\usepackage{booktabs}

\usepackage[T1]{fontenc}
\usepackage[utf8]{inputenc}

\usepackage{microtype}

\usepackage{inconsolata}

\usepackage{graphicx}

\usepackage{listings}
\usepackage{xcolor}
\usepackage{multirow}

\usepackage{amsmath}
\usepackage{enumitem}

\usepackage[most]{tcolorbox} % Load tcolorbox package

\title{CODI: Compressing Chain-of-Thought into Continuous Space via Self-Distillation}

\author{Zhenyi Shen$^{1}$,
        Hanqi Yan$^{1}$,
        Linhai Zhang$^{1}$,
        Zhanghao Hu$^{1}$,
        Yali Du$^{1,2}$,
        Yulan He$^{1,2}$ \\
        $^1$King's College London \hspace{1cm}
        $^2$The Alan Turing Institute \\
        \texttt{\{zhenyi.shen, hanqi.yan, linhai.zhang, zhanghao.hu\}@kcl.ac.uk} \\
        \texttt{\{yali.du, yulan.he\}@kcl.ac.uk}
  }
\usepackage{float}

\begin{document}

\maketitle

\begin{abstract}
%Chain-of-Thought (CoT) enhances Large Language Models (LLMs) by making them reason step-by-step in natural language. However, natural language is shown optimized for communication rather than reasoning, evidenced by its verbosity and redundancy for computational tasks. Prior implicit CoT methods enable LLM to enhance reasoning without natural language, they, however, have consistently underperformed the standard CoT method -- explicit CoT. We propose \textbf{CODI} (Continuous Chain-of-Thought via Self Distillation), a novel training framework that compresses natural language CoT tokens into continuous space. CODI jointly learns a teacher task (explicit CoT) and a student task (implicit CoT), and distills reasoning capability in natural language to continuous space by aligning the hidden activations of a designated token. Our results indicate that CODI is the first implicit CoT method to match explicit CoT’s performance on GSM8k at the GPT-2 scale, achieving 3.1× compression and surpassing the previous state-of-the-art by 28.2\% in accuracy. Notably, CODI retains interpretability by decoding its continuous thoughts, offering certain transparency into its reasoning process. These results demonstrate that LLMs can reason effectively not only by generating more natural language tokens, but also by generating continuous tokens when guided appropriately.

Chain-of-Thought (CoT) reasoning enhances Large Language Models (LLMs) by encouraging step-by-step reasoning in natural language. However, leveraging a latent continuous space for reasoning may offer benefits in terms of both efficiency and robustness. Prior implicit CoT methods attempt to bypass language completely by reasoning in continuous space but have consistently underperformed compared to the standard explicit CoT approach.
We introduce \textbf{CODI} (Continuous Chain-of-Thought via Self-Distillation), a novel training framework that effectively compresses natural language CoT into continuous space. CODI jointly trains a teacher task (Explicit CoT) and a student task (Implicit CoT), distilling the reasoning ability from language into continuous space by aligning the hidden states of a designated token. 
Our experiments show that CODI is the first implicit CoT approach to match the performance of explicit CoT on GSM8k at the GPT-2 scale, achieving a 3.1x compression rate and outperforming the previous state-of-the-art by 28.2\% in accuracy. CODI also demonstrates robustness, generalizable to complex datasets, and interpretability. These results validate that LLMs
can reason effectively not only in natural language, but also in a latent continuous space. Code is available at https://github.com/zhenyi4/codi.

%implicit CoT as not only a more efficient but a powerful alternative to explicit CoT.
%Chain-of-Thought (CoT) enhances Large Language Models (LLMs) by enabling step-by-step reasoning in natural language. However, the language space may be suboptimal for reasoning. While implicit CoT methods attempt to enable reasoning without explicit CoT tokens, they have consistently lagged behind explicit CoT method in task performance. We propose CODI (Continuous Chain-of-Thought via Self-Distillation), a novel framework that distills CoT into a continuous space, where a shared model acts as both teacher and student, jointly learning explicit and implicit CoT while aligning their hidden activation on the token generating the final answer. CODI is the first implicit CoT method to match explicit CoT’s performance on GSM8k, surpassing the previous state-of-the-art by 28.2\% while achieving a 3.1× compression ratio. Furthermore, CODI demonstrates scalability, robustness, and generalizability to more complex CoT datasets. Additionally, CODI retains interpretability by decoding continuous thoughts, making its reasoning process transparent. Our findings establish continuous reasoning as not only a more efficient but a powerful alternative to explicit CoT.
\end{abstract}

\section{Introduction}
Large Language Models (LLMs) have exhibited remarkable reasoning capabilities \cite{gpt4o,anthropic,germini}, with Chain-of-Thought (CoT) \cite{wei2023chainofthoughtpromptingelicitsreasoning} emerging as a key technique for enabling step-by-step reasoning. The success of CoT can be explained as it allows human-like deliberate thinking when computing a sequence of reasoning tokens before deriving the final answer ~\citep{Kahneman2011ThinkingFA}.

%More recently, reasoning in the latent space has become an emergent research direction and it can also enable the model to \textit{adaptively} think more in continuous space.
However, conventional CoT-based methods only rely on natural language tokens as the medium for reasoning. While prior work on prompt learning \cite{lester-etal-2021-power} has demonstrated that transforming discrete prompts into continuous representations can lead to efficient yet effective reasoning \cite{li-liang-2021-prefix}. This motivates us to investigate if CoT reasoning can similarly benefit from continuous representations.
Compared to natural language, reasoning in continuous space offers the following advantages. First, verbalizing the reasoning process can be inefficient, as many tokens are devoted to communication rather than computation \cite{li2024chainthoughtempowerstransformers}. Second, learning annotated CoTs token-by-token may cause models to overfit on superficial linguistic cues \cite{lin2025criticaltokensmattertokenlevel}. While continuous representations—without the need to mimic explicit targets—introduce a softer prior, which may lead to improved robustness.

%which could hinder the model's generalization capability in out-of-distribution scenarios. %As a result, implicit CoT has emerged as a promising research direction \cite{qu2025surveyefficientreasoninglarge,feng2025efficientreasoningmodelssurvey}.

\begin{figure}
\small
\centering
   \includegraphics[width=\columnwidth]{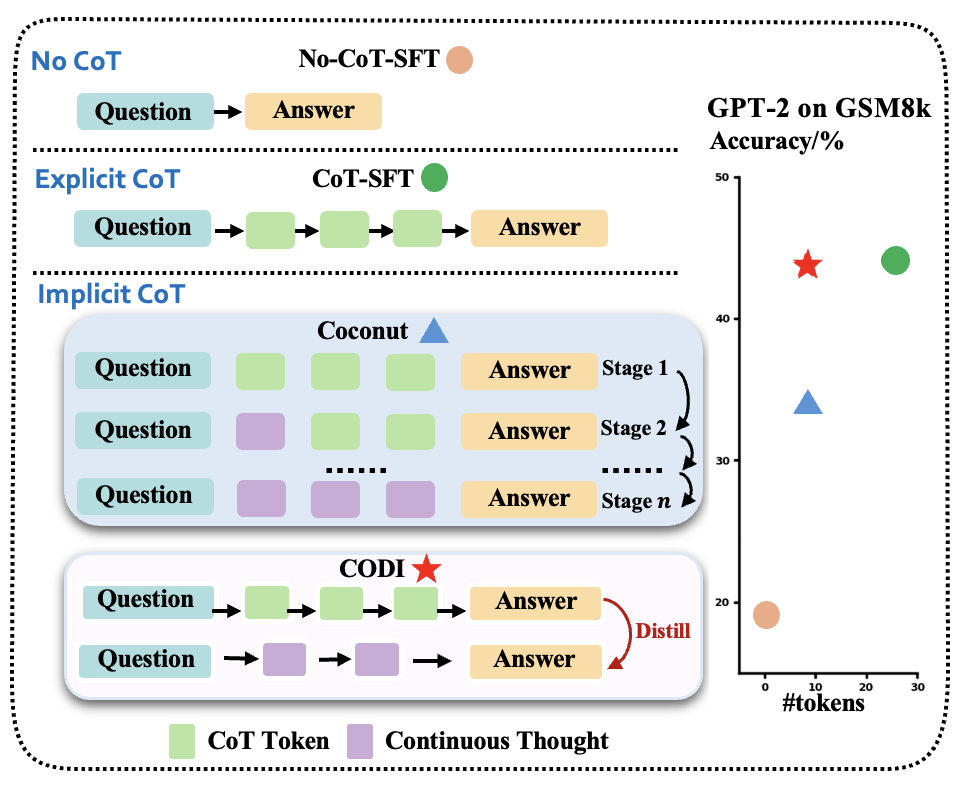}
  \caption{
  %\footnotesize 
  %\hq{CoT tokens should be more than continuous tokens. more simple symbols, like triangle as cot, square as continuous, no need for the lm loss annotation.} 
  Comparison of reasoning strategies. \textbf{No-CoT-SFT}: Train model on (Q,A) pairs via SFT.
  \textbf{CoT-SFT}: Train model on (Q, CoT, A) triples via SFT, i.e., with explicitly annotated CoT reasoning steps. 
  \textbf{Coconut}: requires multi-stage training to progressively replace CoT tokens with continuous representations.
  \textbf{CODI}: achieves this in a single stage by compressing CoT tokens into continuous space via self-distillation.
 % adopt implicit CoT through continuous thought using different training strategies. 
 %Results on GSM8k show that CODI outperforms Coconut and achieves balanced accuracy and token efficiency.
 }
      \vspace{-10pt} % Reduce space below the figure
  \label{fig:codi_illustrate}
  \vspace{-5pt}
\end{figure}

An implicit CoT algorithm replaces natural language tokens with continuous representations for reasoning as shown in Figure \ref{fig:codi_illustrate} (left). To effectively learn these representations, \citet{pfau2024letsthinkdotdot, goyal2024thinkspeaktraininglanguage} pretrain the model with additional thinking tokens from scratch. More recently, the state-of-the-art method, Coconut \cite{Coconut} adopts a curriculum learning strategy \cite{icot_si} that gradually replaces the initial CoT tokens with continuous thoughts. This strategy encourages continuous thoughts to behave like the removed CoT tokens. Although Coconut has greatly improved upon earlier implicit CoT methods in terms of performance ~\citep{goyal2024thinkspeaktraininglanguage,icot_si}, it lags behind \texttt{CoT-SFT} by a large margin as shown in Figure \ref{fig:codi_illustrate} (right). We hypothesize that this performance gap is due to forgetting across stages in the curriculum learning process \cite{vijjini2021analyzingcurriculumlearningsentiment}. This prompts us to ask: \textit{\textbf{Can implicit CoT methods achieve the reasoning capability comparable to \texttt{CoT-SFT} while maintaining their efficiency advantages?}}

%To alleviate the above issues and further enhance the performance of existing implicit CoT, 
To address this, we propose a novel training framework: \textbf{CODI (Continuous Chain-of-Thought via Self Distillation)}. CODI enables implicit CoT learning in a single training step by leveraging self-distillation, thereby avoiding the forgetting issues inherent in curriculum learning. In doing so, it achieves performance comparable to \texttt{CoT-SFT} while being significantly more efficient.
CODI enables implicit CoT reasoning through a joint learning setup involving a \textit{teacher} task and a \textit{student} task. 
The teacher learns from the annotated CoT tokens using a cross-entropy loss, while the student generates a small number of continuous thoughts before producing the final answer, representing implicit CoT reasoning. 
%To encourage flexible thinking, we do not constrain the student's intermediate thoughts to match any specific target. 
We do not constrain the student's continuous thoughts to match any specific target.
%Instead, we injects the teacher's knowledge of reasoning into the student by aligning the hidden activations of a \textit{distilled token}, the token responsible for generating the first answer token, between the teacher and the student.
Instead, we transfer the teacher’s reasoning knowledge to the student through a form of representation alignment at the position of answer generation, where the essence of the reasoning process is captured \cite{orgad2024llmsknowshowintrinsic}.
This allows the student to effectively mimic the teacher's reasoning pattern in continuous space without rigid constraints.
We refer to this mechanism as \textit{self-distillation} \cite{wang2023selfinstructaligninglanguagemodels,Gou_2021}, emphasizing the model’s ability to distill one of its own behaviors into another.

The main contributions are threefold:
\begin{itemize}[leftmargin=*,itemsep=0pt, topsep=0pt,nolistsep]
    \item We propose CODI, a novel self-distillation framework that enables LLMs to reason in a compact continuous space, providing an alternative to accelerate reasoning with high performance.
    \item We demonstrate the effectiveness of distilling knowledge from explicit CoT to implicit CoT by aligning the hidden activations of a single token.
    \item Extensive experiments show that CODI is robust, generalizable to complex CoT datasets, and offers a reasonable level of interpretability. %Additionally, CODI maintains interpretability, making its reasoning process more transparent.
\end{itemize}

\begin{figure*}[t]
  %\vspace{-80pt} % Reduce space above the figure
  \includegraphics[width=\textwidth]{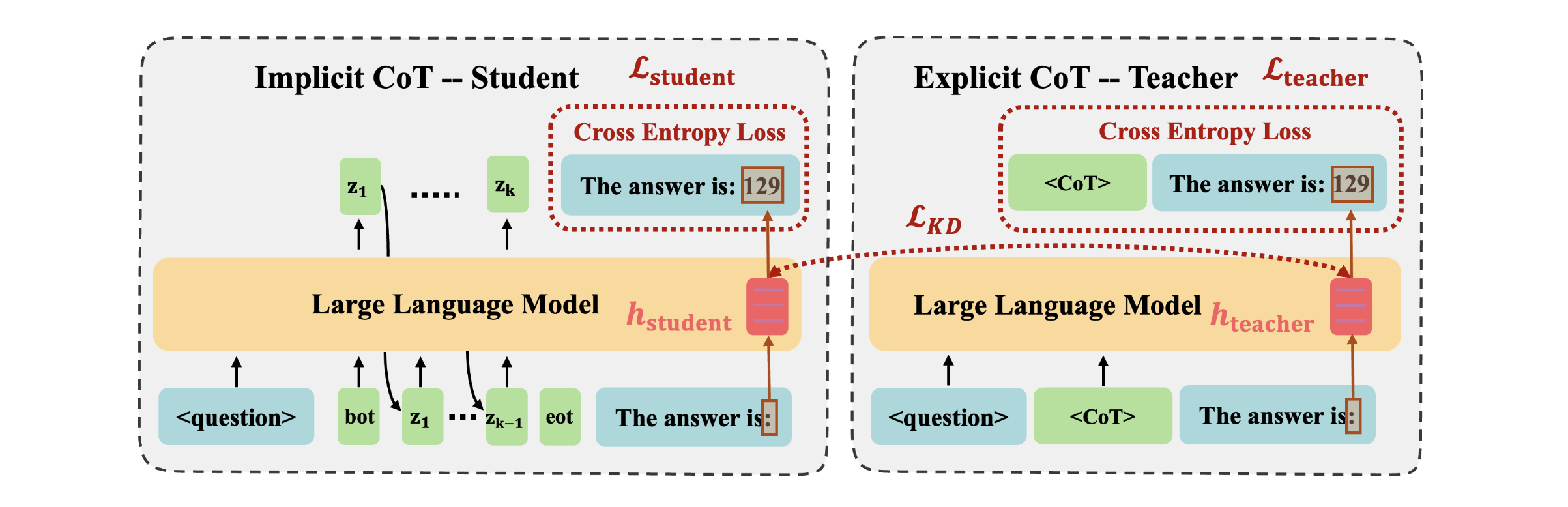}
%\vspace{-105pt} % Reduce space below the figure
%\small
  \caption{%\footnotesize
  \textbf{CODI} enables the model to generate implicit continuous CoTs by jointly training a student task and a teacher task, and distills knowledge from the teacher to the student.
  The \textbf{Student} task (left) generates the answer by autoregressively decoding continuous thoughts starting from a learnable \texttt{bot} token, while the \textbf{Teacher} task (right) generates the answer using the groundtruth CoT via teacher forcing. Both tasks learn the generated texts via cross-entropy loss ($\mathcal{L}_{\text{student}}$ and $\mathcal{L}_{\text{teacher}}$), and share the same LLM. Knowledge distillation is achieved by applying $\mathcal{L}_{\text{KD}}$ (L1 loss) between student and teacher hidden activation across all layers ($\textbf{h}_{\text{student}}$ and $\textbf{h}_{\text{teacher}}$). }
  \label{fig:codi_method}
  \vspace{-5pt}
\end{figure*}
%\hq{add a sentence of existing limitation, based on the limitation, propose CODI}
%Unlike explicit CoT, CODI enables reasoning with significantly fewer tokens in a more compact continuous space.\hq{this is not the key contribution of CODI}.
%\hq{probably, you can highlight that the CODI performance is clearly xx\% better than implcitCoT/CoConut}, surpassing previous state-of-the-art methods by a substantial margin. 
%\hq{personally, I think you should focus more on CODI is better than coconut, less on CODI can compensate the shortcomings of explict CoT. Based on that, you can adjust according the two parts.}
\section{Related Work}
%\hq{usually, the ACL* papers discuss related work intermediately after the intro, and ML papers at the end, but it is not a big deal. While, as you mention the distinguished features of CODI, probably, move to the front can attract more attention. Usually, after experiment, reviewers will not read the paper any more.}
%1. CoT reasoning
%2. latent CoT reasoning
\paragraph{Implicit Chain-of-Thought Reasoning.} 
%Chain-of-Thought (CoT) prompting \cite{wei2023chainofthoughtpromptingelicitsreasoning} significantly improves performance on arithmetic, symbolic, and commonsense reasoning tasks \cite{wei2023chainofthoughtpromptingelicitsreasoning,sprague2024cotcotchainofthoughthelps} via generating explicit intermediate reasoning steps before the final answer.
Implicit CoT methods aim to enhance reasoning without verbalizing intermediate steps as in CoT, thereby accelerating inference speed. Theoretical work \cite{Strobl_2024,merrill2024expressivepowertransformerschain} establishes that additional computational tokens enhance transformers’ reasoning capacity.
Empirical studies \cite{pfau2024letsthinkdotdot,goyal2024thinkspeaktraininglanguage} validate these insights by training LLMs with extra dummy tokens before answering though in a limited scale and effect.
Recent efforts \cite{icot_kd,icot_si} distills CoT reasoning by fine-tuning. They improve over the No-CoT baseline, but fall behind CoT finetuning possibly due to discarding all intermediate tokens.
Addressing this, Coconut \cite{Coconut} reintroduces intermediate reasoning tokens via autoregressive hidden state propagation, combining curriculum learning from \cite{icot_si}. While this achieves some improvement over \cite{icot_si}, Coconut still lags behind explicit CoT, which we attribute to forgetting in curriculum learning.
CODI replaces curriculum learning with a novel self-distillation framework, enabling a single-step learning process that avoids forgetting issues. 
%Unlike \cite{icot_kd}, which aligns hidden states across all CoT tokens, CODI aligns only the hidden states of the final answer-generating token, bypassing the challenge of dynamic CoT lengths.
Our work is also inspired by in-context compression \cite{ge2024incontextautoencodercontextcompression,li2024500xcompressorgeneralizedpromptcompression}, though our work is compressing the generation instead of the existing contexts. Concurrent works \cite{xu2025softcotsoftchainofthoughtefficient,liu2024deliberationlatentspacedifferentiable,su2025tokenassortedmixinglatent} explore latent reasoning, but still rely on explicit CoT generation. Looped transformers \cite{geiping2025scalingtesttimecomputelatent,Saunshi2025ReasoningWL,Yu2025EnhancingAC} also support latent reasoning, though they primarily vary in model depth without introducing. In contrast, CODI emphasizes increasing reasoning capability through additional tokens.

\paragraph{Knowledge Distillation.} Knowledge distillation (KD) \cite{Gou_2021,xu2024surveyknowledgedistillationlarge} has emerged as a key strategy for transferring CoT reasoning capabilities from teacher to student models. Traditional approaches \cite{hsieh2023distillingstepbystepoutperforminglarger,ho2023largelanguagemodelsreasoning} train smaller student models to mimic step-by-step outputs from larger teacher LLMs, motivated by findings that CoT reasoning emerges predominantly in large models \cite{wei2023chainofthoughtpromptingelicitsreasoning}. 
Self-distillation \cite{yang2024selfdistillationbridgesdistributiongap,dong2024undialselfdistillationadjustedlogits} leverage self-distillation to preserve the model's original behavior, akin to the KL divergence loss used in RLHF \cite{ouyang2022traininglanguagemodelsfollow}. 
%Recently, more advanced distillation strategies \cite{wang2023scottselfconsistentchainofthoughtdistillation,li-etal-2024-mode,feng2024keypointbasedprogressivechainofthoughtdistillation} have been proposed to better distill CoTs into student models.
%%%%%%%%%%%commented by hanqi%%%%%%%%
% Our work is, however, closely related to self-distillation, as we use the same model for both the teacher and the student. In contrast, CODI strengthens the teacher by providing it with richer input contexts, enabling the student to learn from it like knowledge distillation.
% Furthermore, while most knowledge distillation methods operate in the response space \cite{hsieh2023distillingstepbystepoutperforminglarger,ho2023largelanguagemodelsreasoning}, CODI extends this by incorporating an L1 loss in the feature space \cite{kd_feature} and carefully selecting the token location to be distilled \cite{orgad2024llmsknowshowintrinsic}. This approach captures the intermediate representations critical for reasoning more freely.
% Finally, CODI distinguishes itself by allowing the student to reason in a space other than natural language, a capability rarely explored in prior knowledge distillation works. This innovation enables more flexible and efficient reasoning, particularly in continuous latent spaces.
%%%%%%%%%%%%replacement from hanqi%%%%%%%
Our work is based on self-distillation framework, but further strengthens the teacher by providing it with richer input contexts, enabling the student to learn from it like knowledge distillation. Since the teacher and student tasks differ, CODI can also be viewed as a form of multitask learning \cite{crawshaw2020multitasklearningdeepneural}.
Moreover, CODI distinguishes itself by allowing reason in the latent space other than natural language, which is rarely explored in prior knowledge distillation works. This innovation enables more flexible and efficient reasoning.
%\hq{probably, need to highlight that why your model also share the similar characteristics of the self-distill. For example, you are distilling from the modal capacity(i,e., you want to keep) in the last stage? to avoid any mis-usage of the word `distillation' for expert reviewer.}
% particularly in continuous latent spaces.
%%%%%%%%%%%%%%%%%%%%%%%%%%%

%\input{sections/method_reorder.tex} 
\section{CODI: Continuous Chain-of-Thought via Self Distillation}
%\hq{teacher is the alignment reference, just briefly intro; then introduce student, two loss, so move distilled part into student task. also, saying in the overview, that student task is the target/resulted model for the following inference.}
%\hq{follow the sketch of intro: teacher task is a CoT; Two objectives of student tasks, (i) answer should be the same (ii) distill knowledge into distilled tokens.}

Unlike traditional CoT reasoning, CODI bypasses autoregression in the vocabulary space, and directly connects the last hidden representation to the subsequent input. 
The key challenge in training such a model with continuous thoughts lies in designing an appropriate training objective. Conventional reasoning learning in explicit CoT fine-tuning relies on a cross-entropy loss over annotated CoT tokens, which inevitably leads to discrete CoT token generation—contradicting the definition of implicit CoT.
\subsection{Overview}
CODI addresses this challenge by introducing a self-distillation framework (Figure \ref{fig:codi_method}) with two training tasks: a teacher task and a student task. The teacher task learns explicit CoT reasoning, while the student task learns implicit CoT reasoning. Knowledge distillation is achieved by aligning the hidden activations of a key token from the teacher to the student via $\mathcal{L}_{KD}$. The overall training objective is a weighted sum of three losses:
\begin{equation}
%\small
\mathcal{L} = \alpha \mathcal{L}_{\text{student}} + \beta \mathcal{L}_{\text{KD}} + \gamma \mathcal{L}_{\text{teacher}},
\label{equ:total_loss}
\end{equation}
where $\alpha$, $\beta$, and $\gamma$ are hyperparameters controlling the balance among the objectives.\footnote{A Python implementation of this framework is provided in Figure~\ref{fig:code}.}

\subsection{Teacher Task}
The teacher task (Figure \ref{fig:codi_method}, right) learns explicit CoT using a cross-entropy loss:
\begin{equation}
%\small
\mathcal{L}_{\text{teacher}} = -\frac{1}{N} \sum_{i=1}^{N} \log P(r_i \mid r_{1:i-1}, Q),
\label{equ:teacher}
\end{equation}
where $P$ denotes the output probability distribution of the LLM, $Q$ represents the question tokens, and $r = \left[ c, y \right]$ is the concatenated sequence of the CoT reasoning tokens $c$ and the final answer token $y$.

\subsection{Student Task}
The student task (Figure \ref{fig:codi_method}, left), which performs implicit CoT reasoning, generates continuous thoughts by autoregressively propagating the last hidden states. This process begins with a learnable \texttt{<bot>} (\textit{begin-of-thoughts}) token and proceeds until a learnable \texttt{<eot>} (\textit{end-of-thoughts}) token is reached. The model then learns the final answer from the \texttt{<eot>} token using a cross-entropy loss:
\begin{equation}
%\small
\mathcal{L}_{\text{student}} = - \frac{1}{N} \sum_{i=1}^{N} \log P(y_i \mid y_{1:i-1}, Q, Z),
\label{equ:student}
\end{equation}
where $y$ denotes the answer label, $Q$ the question tokens, and $Z$ the continuous thoughts.

Additionally, a two-layer MLP followed by layer normalization transforms the hidden representations of continuous thought tokens before feeding them into the next step for the purpose of better discriminating the latent space and the token space.

\subsection{Self-Distillation}
\label{sec:self-distill}
If the model learns only with the student task, it benefits only marginally from the additional computation \cite{goyal2024thinkspeaktraininglanguage} due to the absence of supervision for continuous thoughts.

\paragraph{Distillation in Feature Space.}
To provide explicit supervision to guide continuous thoughts, we adopt a feature-level distillation strategy. Recent work \cite{li2024incontextlearningstatevector, liu2024incontextvectorsmakingcontext} demonstrates that in-context examples influence the final query token by shifting its hidden activation values. Extending this idea, we show that CoT tokens similarly induce a shift in hidden activation values of a query token (can be a probing token like "Answer") compared to a sequence without CoT, as formalized in Equation~\ref{equ:shift}:
\begin{equation}
\label{equ:shift}
    \mathbf{h}^l_{\text{CoT}} \approx \mathbf{h}^l_{\text{no-CoT}} + f\Big(W_V R(W_K R)^T\textbf{q}\Big),
\end{equation}
where $\textbf{q}$ is the query token, $\mathbf{h}^l_{\text{CoT}}$ is the hidden activations at layer $l$ with CoT, $\mathbf{h}^l_{\text{no-CoT}}$ is the corresponding activation without CoT, and the remaining term quantifies the shift introduced by the CoT rationale $R$. A formal proof of this “\textit{CoT shift}” phenomenon is provided in Appendix~\ref{appx:proof}.

This decomposition suggests that the key information from CoT reasoning accessible to the query token is embedded in the shift term $f(\cdot)$. Therefore, by encouraging the student's hidden activations $\mathbf{h}^l_{\text{student}}$ to align with the teacher's $\mathbf{h}^l_{\text{teacher}}$, we are able to transfer the reasoning capability from explicit CoT to implicit CoT.

\paragraph{The Distilled Token.}
Rather than aligning with all tokens in the query sentence, we select a \textit{distillation token} for alignment. Inspired by the recent observations \citep{orgad2024llmsknowshowintrinsic} that the hidden activations of the token intermediately preceding the answer, i.e., the colon (“:”) in the answer prompt “\textit{The answer is:}” (as shown in Figure~\ref{fig:codi_method}), encodes essential reasoning information. We select this token's hidden activations, $\textbf{h}$, for distillation. Alternative answer prompts and distillation tokens are also effective, and the corresponding ablation studies are reported in Appendix \ref{sec:ablation_distill_token}.

\paragraph{Loss Function.}

As a result, we formulate a loss function that aligns the teacher’s and student’s hidden activations across all layers at the selected distillation token for the student's implicit CoT learning. To ensure a one-way flow of knowledge, we apply a stop-gradient operation on $\mathbf{h}^l_{\text{teacher}}$, only allowing the teacher to influence the student:

\begin{equation}
%\small
\mathcal{L}_{\text{KD}} = \frac{1}{M} \sum_{l=1}^M |\text{sg}[\textbf{h}_{\text{teacher}}^l]-\textbf{h}_{\text{student}}^l|,
\label{equ:kd}
\end{equation}
where $M$ indicates the number of layers in the LLM, sg denotes the stop-gradient operation, and $\textbf{h}^l$ is the hidden activations of the LLM's $l$-th layer for the token position corresponding to the colon “:” in our design.

\subsection{Training and Inference}
% \paragraph{Training Details.}
%\hq{personally, I think it is better to move part of the discussion of why joint training and independent training here, (1) the method part is too short (2) the motivation why you use joint training should be discussed in the method, and the results showing your motivation is correct should be in experiment}

%In CODI, a single model is utilized for both the teacher and student tasks. This design is motivated by the dual role of the teacher task. (1) The teacher learns to generate standard CoTs as \textbf{behavior cloning}. (2) The teacher provides hidden activations guided by the groundtruth CoT, enabling the student to generate a sequence of continuous thoughts that align with these activation, thereby achieving \textbf{knowledge distillation}. Both components are critical to the training process. Behavior cloning ensures that the model first masters the generation of standard CoTs, which are then transferred to the continuous space. Meanwhile, knowledge distillation offers richer information than the cross-entropy loss, $\mathcal{L}_{\text{student}}$, guiding the continuous CoT generation to produce a "CoT shift" effect that mirrors the teacher's reasoning process. Experimental results supporting this argument are presented in Section~\ref{sec:ablation}.
\paragraph{Training.}  
The continuous thoughts are generated dynamically during training, as they are not known beforehand. To achieve this, we decode them step by step, with a cache storing previous keys and values to maintain efficiency. When applying a distance metric between two hidden activations, we observed significant norm variations across layers \cite{icot_kd,cheng2024compressedchainthoughtefficient}. To address this, we normalize each layer’s hidden activations by dividing them by the standard deviation of the teacher’s corresponding hidden activations within the current batch.

%For the distillation task, we employed the same model for the teacher task and the student task for two reasons: (1) \textbf{Reference Learning}: The model must first know how to perform explicit CoT, then it can learn the more advanced implicit CoT; (2) \textbf{Training Efficiency}: It is either inconvenient to train a teacher model beforehand or costing certain memory to load two separate models during training. Ablations have been done when exploring how the teacher and student should be trained in Table \ref{tab:ablations}.

For the distillation task, we adopt the same model for both the teacher and student roles for two primary reasons. (1) \textbf{Reference Learning:} The model must first learn to perform explicit CoT reasoning before it can effectively compress and transfer this capability into continuous space as implicit CoT. (2) \textbf{Training Efficiency:} While it is feasible to train separate teacher and student models—as explored in Section~\ref{sec:ablations}—this setup introduces additional complexity. The teacher must be pre-trained, and maintaining two distinct models during training doubles memory consumption.

%(1) \textbf{Warm-up}: When the teacher trains alongside the student, it creates a warm-up effect for $\mathcal{L}_{\text{KD}}$. Both components start from the same initialization point, diverge during training, and gradually converge as the student adapts. In contrast, a static pre-trained teacher initially presents an overly challenging objective, as its hidden states reflect fully developed reasoning patterns that the untrained student cannot immediately match. (2) \textbf{Shared model representations}: Using the same model mitigates alignment issues in hidden activations that arise when using separate models, enabling smoother and more effective information transfer between the teacher and student. The corresponding ablation studies, which validate these findings, are detailed in Table \ref{tab:ablations}.

For training data, we exclude the final CoT step—the step responsible for generating the final answer—because including this step could allow the teacher’s hidden activations to take a shortcut. Specifically, the model might directly copy the result from the last CoT step to the token responsible for generating the exact answer token, bypassing the reasoning process. This behavior would undermine the quality of the target hidden activations, as they would no longer fully encode the reasoning patterns. The ablation results demonstrating the impact of this exclusion are presented in Table \ref{tab:ablations}.

%discovered that excluding the final CoT step—the step responsible for generating the final answer—leads to improved performance. We hypothesize that including this step might allow the teacher’s hidden activations to take a shortcut, directly copying the result from the last step to the token responsible for generating the exact answer token. This behavior undermines the target hidden activations. 

%By excluding the final step, our method performs a two-stage reasoning process, as indicated in Equation \ref{equ:two_step_reason}. In this configuration, the model first predicts the second-to-last intermediate result and then infers the final answer, as shown in Equation \ref{equ:two_step_reason}. This two-stage reasoning approach has also been demonstrated to be effective in \cite{zhang2024multimodal}.

%4. Inference process
\paragraph{Inference.} The inference process in CODI mirrors the student task during training (Figure \ref{fig:codi_method}, left). The model autoregressively decodes $n$ continuous thoughts following the question and the \texttt{bot} token. Once the reasoning process is complete, the \texttt{eot} token is manually inserted to terminate continuous reasoning and switch the model to language generation mode, decoding the final answer.

\begin{figure*}[t]
  \includegraphics[width=\textwidth]{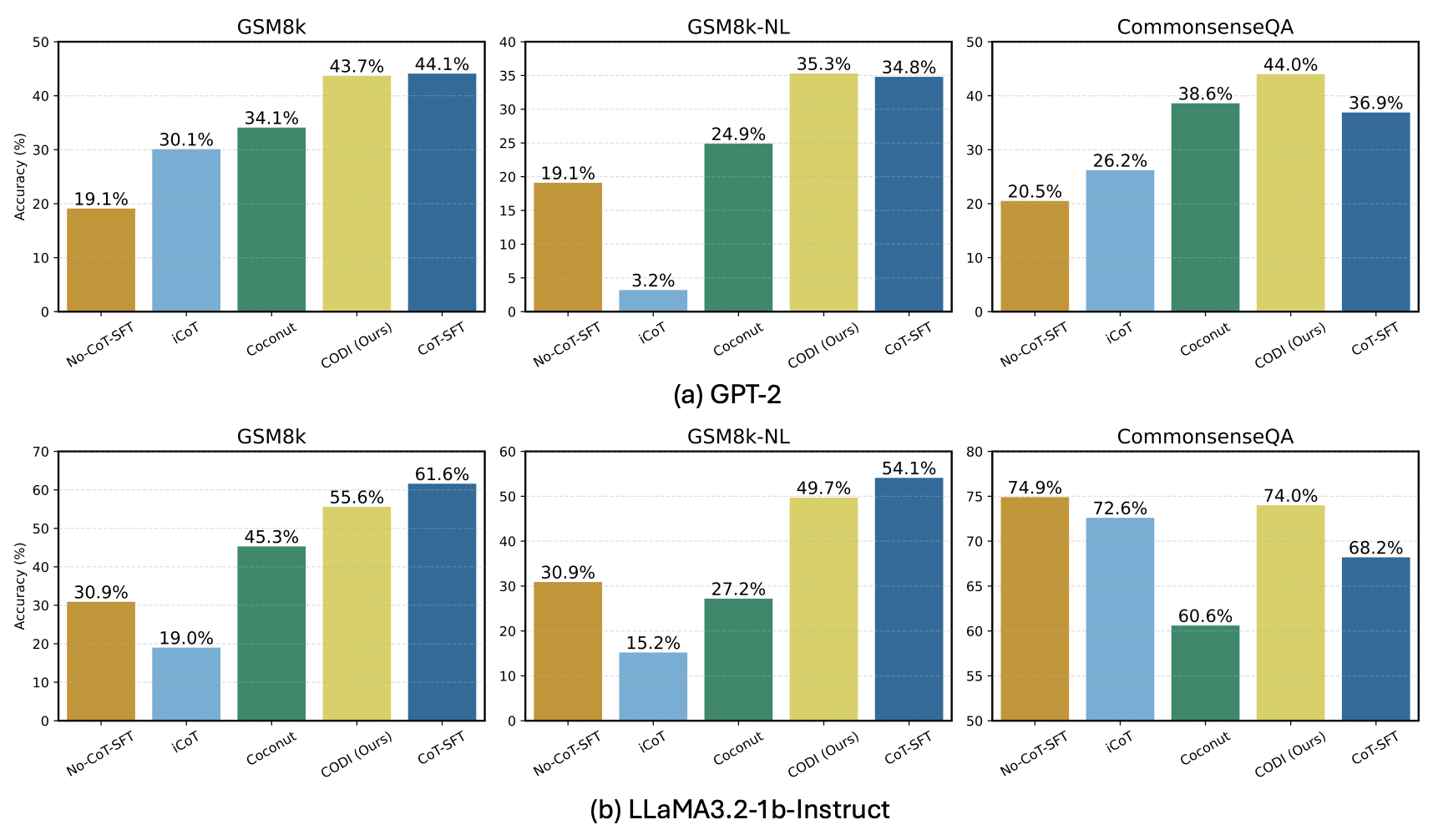}
\vspace{-15pt} % Reduce space below the figure
%\small
  \caption{Results on five datasets (\textbf{Top}: GPT-2, \textbf{Bottom}: LLaMa3.2-1b-Instruct). % GSM8k, GSM8k-NL, SVAMP, GSM-Hard, and MultiArith. 
  CODI consistently outperforms all previous implicit CoT methods by a substantial margin. When using GPT-2, CODI even matches the performance of CoT-SFT on the in-domain GSM8k and GSM8k-NL datasets.} % and surpasses it on out-of-distribution benchmarks.}
  \label{fig:codi_result}
  \vspace{-10pt}
\end{figure*}

\section{Experiments}
%We validate the feasibility of LLM reasoning with latent CoT through experiments on five datasets. We mainly compare the model generated answers with the ground truth.
%\hq{add an overview, what aspects are you going to evaluate the model? Or you can move your RQX here}
%\hq{the caption of table/figures should briefly include your conclusion. }
%\hq{delete the existing research questions. Instead, change the interpretability section into further analysis. for each paragraph, explore one point, using that point as the summary sentence at the beginning your paragraph.}

We demonstrate CODI’s effectiveness in continuous space reasoning through experiments on mathematical and commonsense reasoning tasks. %We also evaluate on commonsense reasoning tasks, but find that even fine-tuning on CoT data does not consistently outperform training without it—consistent with \citet{sprague2024cotcotchainofthoughthelps}. Since implicit CoT methods aim to distill CoT signals that offers no benefits in this setting, the results do not truly reflect the effectiveness of the methods. See Appendix~\ref{sec:commonsense} for details.

%As such, implicit CoT is unnecessary for these benchmarks, since it attempts to distill CoT signals that do not improve performance. See Appendix~\ref{sec:commonsense} for details.

%More specifically, we present a series of experiments to address the following Research Questions (\textbf{RQs}).
%\begin{itemize}[leftmargin=*,labelsep=0.5em]
%    \item \textbf{RQ1}: How does CODI compare with other baselines in terms of performance %and efficiency?
%    \item \textbf{RQ2}: How to understand each component in CODI's framework?
%    \item \textbf{RQ3}: Does CODI's continuous thought maintain the interpretability of %standard CoTs?
%\end{itemize}

\subsection{Experimental Setup}
%\hq{it is not common that you discuss the training and evaluation benchmark, separately. you can separate the dataset discussion in in-domain(train=evaluate) and out-of-domain(only dicuss the evaluation dataset is different). Also, use itemize/bullet could be more clearer.}
%We evaluate our proposed CODI primarily on mathematical reasoning tasks, as they are inherently complex and offer abundant CoT training data for fine-tuning.
%Following \cite{Coconut}, we experimented with ProntoQA \cite{saparov2023language} for comparison, but observed that all methods achieved 100\% accuracy, making it less informative for differentiation. Additional details can be found in the Appendix.
%We evaluate our proposed CODI focusing on math reasoning because math reasoning is complex and only in math reasoning, there are abundant CoT training data for finetuning. Besides math, we also evaluate our method on StrategyQA to show it is possible to solve tasks outside math. According to \cite{Coconut}, we also tried ProntoQA \cite{saparov2023language} to compare methods, but we witness all methods are able to achieve 100\%. We leave additional information in Appendix.

%1. Training data. We used Deng's augmented GSM8k for training. We evaluate all methods on 4 math benchmarks. GSM8k is in-domain test set
\paragraph{Training Data.}
We utilize three datasets to train our models--GSM8k-Aug, GSM8k-Aug-NL, and CommonsenseQA-CoT. (1) We use the \textbf{GSM8k-Aug} dataset from \cite{icot_kd}, which has proven effective for training implicit CoT methods \cite{icot_si,Coconut}. This dataset extends the original GSM8k training set \cite{cobbe2021trainingverifierssolvemath} to 385k samples by prompting GPT-4. To facilitate implicit CoT training, all natural language interleaving within the CoT is removed, leaving only structured mathematical expressions such as “\texttt{$<<10\div 5=2>> <<2\times 2=4>> <<6\times 4=24>>$}”. (2) We also use \textbf{GSM8k-Aug-NL}, a version that preserves natural language explanations, to assess both the generalizability and effectiveness of our approach to compress more verbose CoTs. (3) \textbf{CommonsenseQA-CoT} is derived from CommonsenseQA \cite{talmor-etal-2019-commonsenseqa}, a multiple-choice QA dataset built from ConceptNet-based questions \cite{speer2017conceptnet}. As it lacks CoT annotations, we generate 8.1k CoT examples using GPT-4o-mini, filtered by correctness. The 1.2k-example validation set is used for evaluation. Examples and statistics are in Appendix \ref{sec:data_appendix}.
%For commonsense QA, we fine-tune using the StrategyQA training split \cite{geva2021didaristotleuselaptop}, which contains 1.6k samples. Since StrategyQA does not explicitly provide CoT reasoning, we use the "facts" column as the CoT process to assist reasoning.
%\noindent \textbf{Training Data.} For math reasoning, we use the GSM8k-Aug from \cite{icot_kd} for training, which is shown effective for training a continuous or implicit CoT method\cite{icot_si,Coconut}. This dataset is augmented to a size of 378k by prompting GPT-4 with GSM8k's original training data\cite{cobbe2021trainingverifierssolvemath}. In order to facilitate continuous CoT training, they remove all natural languages in the CoT but expressions like "<<10/5=2>> <<2*2=4>> <<6*4=24>>". For commonsense QA, we use StrategyQA's training split \cite{geva2021didaristotleuselaptop} of 1.6k data for finetuning. As StrategyQA does not explicitly define the CoT, we use the facts column as CoT process to aid reasoning.

%2. Evaluations Benchmarks. WE used 3 more math benchmarks as OOD test sets to test robustness of our method.
\paragraph{Evaluation Benchmarks for OOD.} For mathematical reasoning, we assess model robustness on three out-of-domain (OOD) benchmarks: (1) \textbf{SVAMP} \cite{patel-etal-2021-nlp}, a dataset of grade-school arithmetic word problems with simple variations designed for robustness test; (2) \textbf{GSM-HARD} \cite{gao2022pal}, a modified version of the GSM8k test split where numbers are replaced with values of larger magnitude to increase difficulty; and (3) \textbf{MultiArith} \cite{roy2016solvinggeneralarithmeticword}, a subset of MAWPS \cite{koncel-kedziorski-etal-2016-mawps} containing multi-step mathematical word problems. Examples and statistics are in Appendix \ref{sec:data_appendix}.
%SVAMP \cite{patel-etal-2021-nlp}, GSM-HARD \cite{gao2022pal}, and MultiArith \cite{roy2016solvinggeneralarithmeticword}. 

%For commonsense QA, we evaluate the model on the test split of StrategyQA, which consists of 687 examples.
%\noindent \textbf{Evaluations Benchmarks.} For math reasoning, we use the test split of GSM8k\cite{cobbe2021trainingverifierssolvemath} to test model's in-domain math reasoning ability. Moreover, we also test model's robustness by testing them on three OOD benchmarks: SVAMP \cite{patel-etal-2021-nlp}, GSM-HARD \cite{gao2022pal}, MultiArith \cite{roy2016solvinggeneralarithmeticword}. For commonsense QA, we test models on StrategyQA's test split, containing 687 data.

\paragraph{Baselines.} 
We consider the following baselines:  
(1) \textbf{CoT-SFT:} Finetunes the model on CoT data, enabling it to generate intermediate steps followed by the final answer. 
(2) \textbf{No-CoT-SFT:} Finetunes the model using only direct answers, without generating intermediate steps.  
(3) \textbf{iCoT} \cite{icot_si}: Implements a curriculum learning strategy called "Stepwise Internalization", which injects CoT's reasoning patterns into the model's internal states. This allows the model to generate direct answers with higher accuracy during inference.  
(4) \textbf{Coconut} \cite{Coconut}: Build upon iCoT by autoregressively generating intermediate continuous CoT representations, similar to the approach in our work. (5) \textbf{CODI}: our method trained with six continuous thought tokens, matching the setup in Coconut. Baseline (1) is sampled 10 times and their average is reported (temperature=0.1), while baselines (2)–(5) are deterministic models, and their results are reported from a single run. Two base models are considered: GPT-2 \cite{Radford2019LanguageMA} and LLaMA3.2-1b-Instruct \cite{grattafiori2024llama3herdmodels}. More implementation details are in Appendix \ref{sec:hyperparams}.
%\subsection{Baselines} We consider the below baselines. (1) CoT-SFT: finetune the model with CoT data and then generate the final answer. (2) No-CoT-SFT: finetune the model only with the direct answer, no intermediate steps are generated. (3) iCoT \cite{icot_si}: use a curriculum learning strategy called "Stepwise Internalization" to inject CoT into the model's internal activation, and enable the model to generate direct answers during inference time with a higher accuracy. (4) Coconut \cite{Coconut}: use the same training strategy as iCoT but, instead of only generating final answers, generate intermediate continuous CoT autoregressively as that in our work.

\subsection{Main Results}

\paragraph{Mathematical Reasoning.}  
From the results on GSM8k in Figure~\ref{fig:codi_result} (leftmost column), we observe that % presents the results on GSM8k, demonstrating that 
CODI largely outperforms existing implicit CoT methods. With both GPT-2 and LLaMA-1b, CODI surpasses Coconut by over 20\%. Remarkably, CODI is the first continuous CoT method to achieve performance comparable to CoT-SFT when using GPT-2, reaching 99\% of its accuracy. In contrast to iCoT, which fails to scale effectively to larger models, CODI successfully extends to LLaMA-1b, achieving 90\% of CoT-SFT performance. These results verify CODI’s effectiveness on in-domain mathematical reasoning tasks.

\begin{figure}
\small
%\vspace{-5pt}
\centering
  \includegraphics[width=\columnwidth, height=0.5\columnwidth, keepaspectratio]{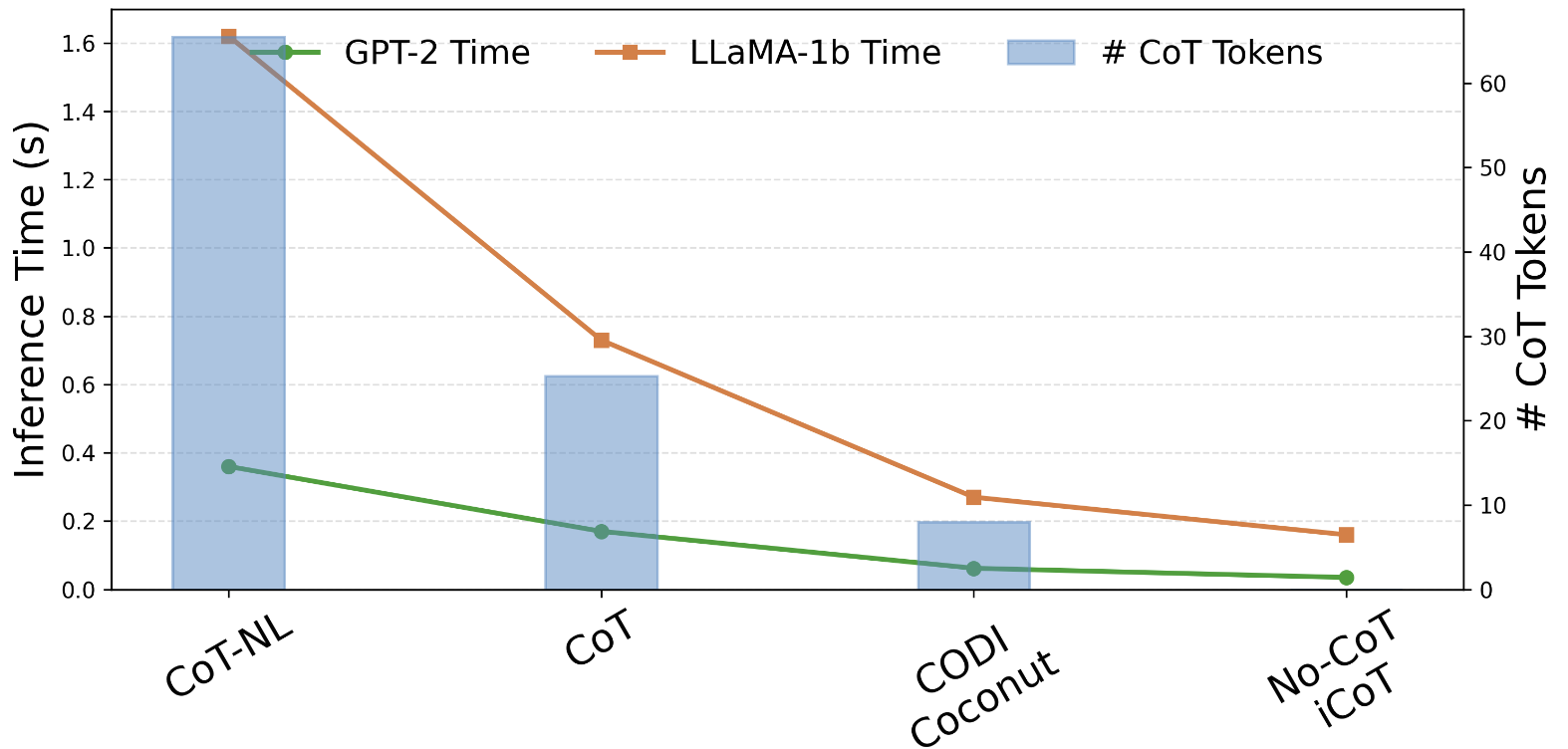}
  \centering
  %\vspace{-10pt}
  \caption{%\footnotesize 
  Efficiency comparison of different reasoning methods in terms of inference time per math problem on GSM8k. Measured with batch size = 1 on an Nvidia A100 GPU. CoT Token counts are shown in parentheses.}
  \label{fig:efficiency}
  %\vspace{-15pt}
\end{figure}

\paragraph{Compress More Verbose CoTs.}
Previous works \cite{icot_si,Coconut} primarily trained on GSM8k-Aug, which consists only of mathematical expressions. 
To evaluate CODI's generalizability, we extend our analysis to a more complex CoT dataset, GSM8k-Aug-NL.
Figure~\ref{fig:codi_result} (2nd column) shows that both GPT-2 and LLaMA-1b perform worse on it compared to GSM8k-Aug. This decrease in performance stems from the additional natural language tokens, which add noise and make imitation learning more difficult. 
Surprisingly, CODI surpasses CoT-SFT when using GPT-2 and achieves a higher relative score improvement on LLaMA1b compared to models trained on GSM8k-Aug. Moreover, CODI surpasses all other implicit CoT methods, especially at the size of LLaMA-1b, suggesting the effectiveness of self-distillation.
Furthermore, with the average CoT length increased to 65.5 (Figure \ref{fig:efficiency}), CODI achieves a compression ratio of 8.2, suggesting that the optimal compression ratio is dataset-dependent. These results demonstrate CODI’s ability to handle more complex CoT training data, showcasing its applicability to diverse reasoning datasets.

\paragraph{Commonsense Reasoning.}
%As shown in Figure~\ref{fig:codi_result} (rightmost column), CoT-SFT underperforms No-CoT-SFT in three out of four settings across two benchmarks and model sizes. This aligns with findings from \citet{sprague2024cotcotchainofthoughthelps}, which show that commonsense reasoning tasks gain little from CoT prompting. Despite this, CODI performs strongly—achieving the best results in two cases and second-best in the remaining two.

As shown in Figure~\ref{fig:codi_result} (rightmost column), CoT-SFT largely outperforms No-CoT-SFT for GPT-2, which performs nearly random guessing (five choices per question). This indicates that training on CoT benefits GPT-2. Interestingly, CODI surpasses even CoT-SFT. We attribute this to GPT-2’s limited capacity for generating coherent natural language CoTs—CoT-SFT struggles to replicate the quality of the training CoTs, whereas CODI faces less burden by reasoning in a continuous space with fewer tokens. For LLaMA-1b, we observe that CoT data actually hurts performance. We think it is because we force the model to reason in GPT-4o-mini’s pattern which may diverge from LLaMA’s original pattern. Interestingly, CODI outperforms CoT-SFT by a large margin and achieves accuracy comparable to No-CoT-SFT. This shows that our latent reasoning model could better capture intermediate thought processes in continuous spaces, demonstrating the benefit of learning latent representations rather than overfitting of specific CoT patterns.

\paragraph{Efficiency.}  
CODI utilizes a fixed set of \textbf{six} continuous thoughts, enclosed by two special tokens, resulting in a total of \textbf{eight} "tokens" for reasoning. As shown in Figure~\ref{fig:efficiency}, CODI achieves substantial efficiency gains, with a speedup of approximately 2.7× (3.1× CoT compression) for compact CoTs trained on GSM8k-Aug and 5.9× (8.2× CoT compression) for verbose CoTs trained on GSM8k-Aug-NL, demonstrating CODI’s effectiveness in reducing reasoning overhead.

%\begin{table}[h]
%    \centering
%    \small
%    \resizebox{1,0\columnwidth}{!}{
%  \begin{tabular}{l|l|l}
%    \toprule
%    \textbf{Method} & GSM8k-Aug & GSM8k-Aug-NL  \\
%                     & Time (\#Tokens) &  Time (\#Tokens) \\
%    \midrule
%    \textbf{GPT-2} && \\
%    CoT-SFT & 0.17s (25.1) & 0.36s (62.1)  \\
%    No-CoT-SFT/iCoT & 0.035s (0) & 0.035s (0)\\
%    CODI/Coconut & 0.062s (8) & 0.062s (8) \\
%    \midrule
%    \textbf{LLaMA-1b} &&\\
%    CoT-SFT & 0.73s (25.4) & 1.62s (68.8)  \\
%    No-CoT-SFT/iCoT & 0.16s (0) & 0.16s (0)\\
%    CODI/Coconut & 0.27s (8) & 0.27s (8) \\
%    \bottomrule
%  \end{tabular}
%  }
%  \caption{Efficiency comparison of different reasoning methods in terms of inference time per math problem on GSM8k. Measured with batch size = 1 on an Nvidia A100 GPU. CoT Token counts are shown in parentheses.}
%  \label{tab:efficiency}
%  \vspace{-15pt}
%\end{table}

\begin{figure}
%\small
%\vspace{-5pt}
\centering
  \includegraphics[width=\columnwidth, height=0.5\columnwidth, keepaspectratio]{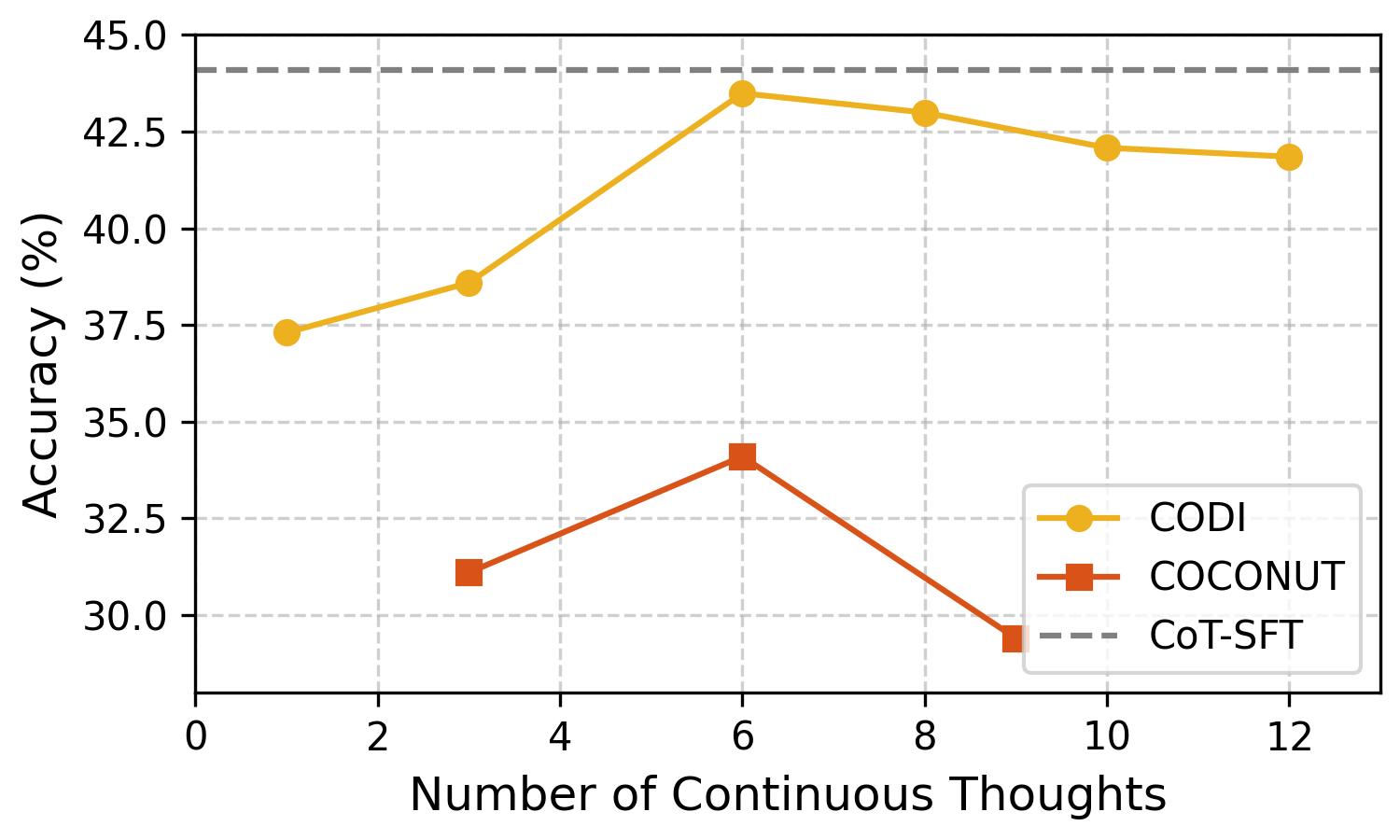}
  \centering
  \caption{%\footnotesize 
  Accuracy on GSM8k against the number of continuous thought tokens used during training.}
  \label{fig:lat_num}
  \vspace{-10pt}
\end{figure}

\paragraph{Compression Ratio.}
The number of continuous thoughts used during training is a crucial hyperparameter, affecting both the computation allocation and the compression ratio. As shown in Figure~\ref{fig:lat_num}, CODI consistently outperforms Coconut across all compression ratios. Interestingly, both methods exhibit a similar trend: accuracy peaks when using six continuous thoughts. We attribute this to the dataset’s structure, specifically the average number of CoT steps.
%We attribute this to the dataset’s structure, where the average number of CoT steps is 2.89.
%—implying that six continuous thoughts approximate two continuous thoughts per step, aligning with CODI's observed computation allocation as shown in Figure\ref{fig:codi_interpretability}. 
When fewer than six continuous thoughts are used, the model lacks sufficient expressiveness to capture reasoning steps effectively. Conversely, beyond six, the additional complexity may not provide further benefits, as most problems do not require additional reasoning steps. Instead, the increased sequence length introduces optimization challenges, outweighing any potential gains.

%\hq{make your figure more flat, it occupies too much space in the vertical direction.}

%\begin{figure}[t]
%  \includegraphics[width=\columnwidth]{figures/mse_visualisation.png}
%  \caption{L1 loss visualisation of number of thoughts used during inference. The model is trained with 6 continuous thoughts.}
%  \label{fig:mse_visualisation}
%\end{figure}

\subsection{Out-of-Distribution (OOD) Evaluation}  
To assess robustness, we evaluate CODI—trained on GSM8k-Aug—on OOD datasets. Remarkably, CODI consistently outperforms all the other implicit CoT baselines and even CoT-SFT across all three OOD benchmarks with GPT-2 %, despite slightly underperforming on the in-domain GSM8k benchmark 
(Table \ref{tab:ood_ratio}). Using LLaMA-1b, CODI also performs better compared to iCoT and Coconut. It also demonstrates stronger performance relative to its in-domain results. %, as reflected in the ratios shown in Table~\ref{tab:ood_ratio}.
We attribute CODI’s robustness to its reduced tendency to overfit. Unlike CoT-SFT, which is trained to mimic exact natural language CoT annotations, CODI generates continuous thoughts without direct imitation targets. This lack of rigid supervision likely prevents memorization and promotes greater adaptability to unfamiliar inputs. %Supporting this hypothesis, CODI exhibits lower training accuracy than CoT-SFT (see Table~\ref{tab:train_acc}), suggesting better generalization.

%the more compact CoT representation in CODI, which makes it more resilient to distribution shifts. 
% \begin{table}[h]
% \centering
% \small
% \resizebox{1.0\columnwidth}{!}{
%     \begin{tabular}{lcccc}
%     \toprule
%     \textbf{Models} & \textbf{GSM8k} & \textbf{SVAMP} & \textbf{GSM-Hard} & \textbf{MultiA} \\
%     \midrule
%     \multicolumn{5}{l}{\textbf{GPT-2}} \\
%     No-CoT-SFT  & 19.1\%  & 16.4\%  & 4.3\% & 41.1\% \\
%     CoT-SFT  & 44.1\% & 41.8\% & 9.8\%  & 90.7\% \\
%     iCoT & 30.1\%  &  29.4\%  & 5.7\% & 55.5\% \\
%     Coconut & 34.1\% & 36.4\%  & 7.9\% & 82.2\% \\
%     CODI     & 43.7\% & 42.9\% & 9.9\%  & 92.8\% \\
%     Ratio    & \underline{0.99} & \textbf{1.03} & \textbf{1.01} & \textbf{1.02} \\
%     \midrule
%     \multicolumn{5}{l}{\textbf{LLaMA-1b}} \\
%     No-CoT-SFT  & 30.9\%  & 44.1\% & 7.1\% & 70.9\%  \\
%     CoT-SFT  & 61.6\% & 66.7\% & 15.6\% & 99.3\% \\
%     iCoT & 19.0\% & 40.9\% & 4.4\% & 39.0\% \\
%     Coconut & 45.3\% & 48.8\% &  9.9\%  & 90.1\%\\
%     CODI     & 55.6\% & 61.1\% & 12.8\% & 96.1\% \\
%     Ratio    & \underline{0.90} & \textbf{0.92} & 0.82 & \textbf{0.97} \\
%     \bottomrule
%     \end{tabular}
% }
% \caption{\footnotesize Performance comparison %between CODI and CoT-SFT 
% on OOD datasets, i.e., trained on GSM8k and evaluated on other datasets. %mathematical reasoning benchmarks. 
% Ratio indicates CODI / CoT-SFT.}
% \label{tab:ood_ratio}
% \vspace{-5pt}
% \end{table}

\begin{table}[h]
\centering
\small
\resizebox{1.0\columnwidth}{!}{
    \begin{tabular}{lccc}
    \toprule
    \textbf{Models} & \textbf{SVAMP} & \textbf{GSM-Hard} & \textbf{MultiA} \\
    \midrule
    \multicolumn{4}{c}{\textbf{GPT-2}} \\
    \midrule
    No-CoT-SFT    & 16.4  & 4.3 & 41.1 \\
    CoT-SFT   & \underline{41.8} & \underline{9.8}  & \underline{90.7} \\
    %\midrule
    iCoT  &  29.4  & 5.7 & 55.5 \\
    Coconut  & 36.4  & 7.9 & 82.2 \\
    CODI    & \textbf{42.9} & \textbf{9.9}  & \textbf{92.8} \\
   % Ratio   & \textbf{1.03} & \textbf{1.01} & \textbf{1.02} \\
    \midrule
    \multicolumn{4}{c}{\textbf{LLaMA-1b}} \\
    \midrule
    No-CoT-SFT   & 44.1 & 7.1 & 70.9  \\
    CoT-SFT   & \textbf{66.7} & \textbf{15.6} & \textbf{99.3} \\
      %\midrule
    iCoT  & 40.9 & 4.4 & 39.0 \\
    Coconut & 48.8 &  9.9  & 90.1\\
    CODI      & \underline{61.1} & \underline{12.8} & \underline{96.1} \\
    %Ratio    & \textbf{0.92} & 0.82 & \textbf{0.97} \\
    \bottomrule
    \end{tabular}
}
\caption{%\footnotesize 
Performance comparison (accuracy \%) %between CODI and CoT-SFT 
on OOD datasets, i.e., trained on GSM8k-Aug and evaluated on other datasets. %mathematical reasoning benchmarks. Ratio indicates CODI / CoT-SFT.
The best results are in \textbf{bold}, and the second-best results are \underline{underlined}.
}
\label{tab:ood_ratio}
%\vspace{-5pt}
\end{table}

%\vspace{-5pt}
\subsection{Ablation Studies} 
\label{sec:ablations}
\begin{table}[h]
%\vspace{-5pt}
    \centering
    %\small
    {!}{
  \begin{tabular}{l|c}
    \toprule
    \textbf{Methods} (GPT-2) & \textbf{Accuracy}  \\
    \midrule
    No-CoT-SFT & 19.1\% \\
    CODI  & 43.7\%  \\
    - separate static teacher & 27.1\% \\
    %\hspace{20pt} w. init student & (wait) \\
    \hspace{20pt} w/ multitask student & 42.2\% \\
    %- ind. trained teacher & $-$ \\
    %\hspace{20pt} w/ multitask student & 42.7\% \\
    - w/o L1 loss & 24.5\% \\
    - w/ CoT last step & 31.7\% \\
    - w/o Projection & 42.5\% \\
    \bottomrule
  \end{tabular}
  }
  \caption{%\footnotesize 
  Ablation studies. \textit{ind. static teacher} refers to introducing an independently trained teacher model. 
  %\textit{w. init student} extends \textit{ind. static teacher} by initializing the student model the same as the teacher model. 
  \textit{w/ multitask student} allows the student model to also learn CoT generation. %\textit{ind. trained teacher} refers to training an independent teacher model along with the student model.
  %\textit{w.o l1 loss} removes the L1 loss constraint between the teacher and student processes.
  %\textit{w. last step} retains the last CoT step in training data.
  %\textit{w.o projection} removes the 2-layer MLP layer.
  }
  \label{tab:ablations}
  \vspace{-10pt}
\end{table}

%%We conduct ablation studies to examine the contribution of each component in CODI (Table \ref{tab:ablations}). 
\paragraph{Independent Teacher.} To evaluate the need of self-distillation, we tested settings where the student does not share the model with the teacher. As observed from Table \ref{tab:ablations}, without learning explicit CoT generation (\texttt{separate static teacher}), the model performs badly and fails to generate meaningful continuous CoTs after decoding. Adding an explicit CoT generation objective (\texttt{w/ multitask student}) significantly restores performance, indicating the importance of \textit{reference learning}.
%Additionally, training the teacher alongside the student (\texttt{ind. trained teacher}) leads to better results than using a pre-trained, static teacher (\texttt{ind. static teacher}), supporting the argument of the warm-up effect.
%Finally, using a unified model (\texttt{CODI}) outperforms maintaining separate teacher-student models (\texttt{ind. trained teacher}), reinforcing the idea that shared model representations help mitigate alignment issues in hidden states.

%\paragraph{Behavior Cloning.} Table~\ref{tab:ablations} shows that when we separate the teacher and the student into two independent models (\texttt{Ind. Teacher}), accuracy drops significantly. However, if the student was still trained with the CoT generation objective while being independent from the teacher (\texttt{Ind. Teacher w/ Multitask}), accuracy remains comparable to the baseline. This suggests that the model must first acquire verbalized CoT generation skills before transferring them to continuous CoT reasoning.

\paragraph{Distillation Loss.} Table~\ref{tab:ablations} also shows that removing the L1 loss (Equation \ref{equ:kd}) linking the teacher and student tasks (\texttt{w/o L1 Loss}) leads to a significant performance drop, indicating the importance of supervision from distillation. While the model performs well in CoT generation due to multitask learning, it fails to integrate this skill into continuous CoT reasoning, treating them as independent tasks rather than a unified reasoning process.

\paragraph{Others.} Keeping the last step of the CoT chain appears to negatively impact performance, supporting our claim that it provides shortcuts. 
The projection layer of continuous thought tokens slightly enhances CODI’s effectiveness. Additional ablations on hyperparameters and the choice of distillation token are reported in Appendix \ref{sec:ablation_hyperparameters} and \ref{sec:ablation_distill_token}. %We also report ablations on the choice of distillation token and hyperparameters in the Appendix \ref{sec:ablation_hyperparameters} and \ref{sec:ablation_distill_token}.

\section{Further Analysis}
We observe that CODI's continuous thoughts exhibit a degree of interpretability. Notably, these patterns cannot not be trivially learned through standard token-by-token fine-tuning (see Appendix~\ref{sec:pattern}).

%\subsection{Interpretability}
\begin{figure*}[t]
\small
  \includegraphics[width=\textwidth, keepaspectratio]{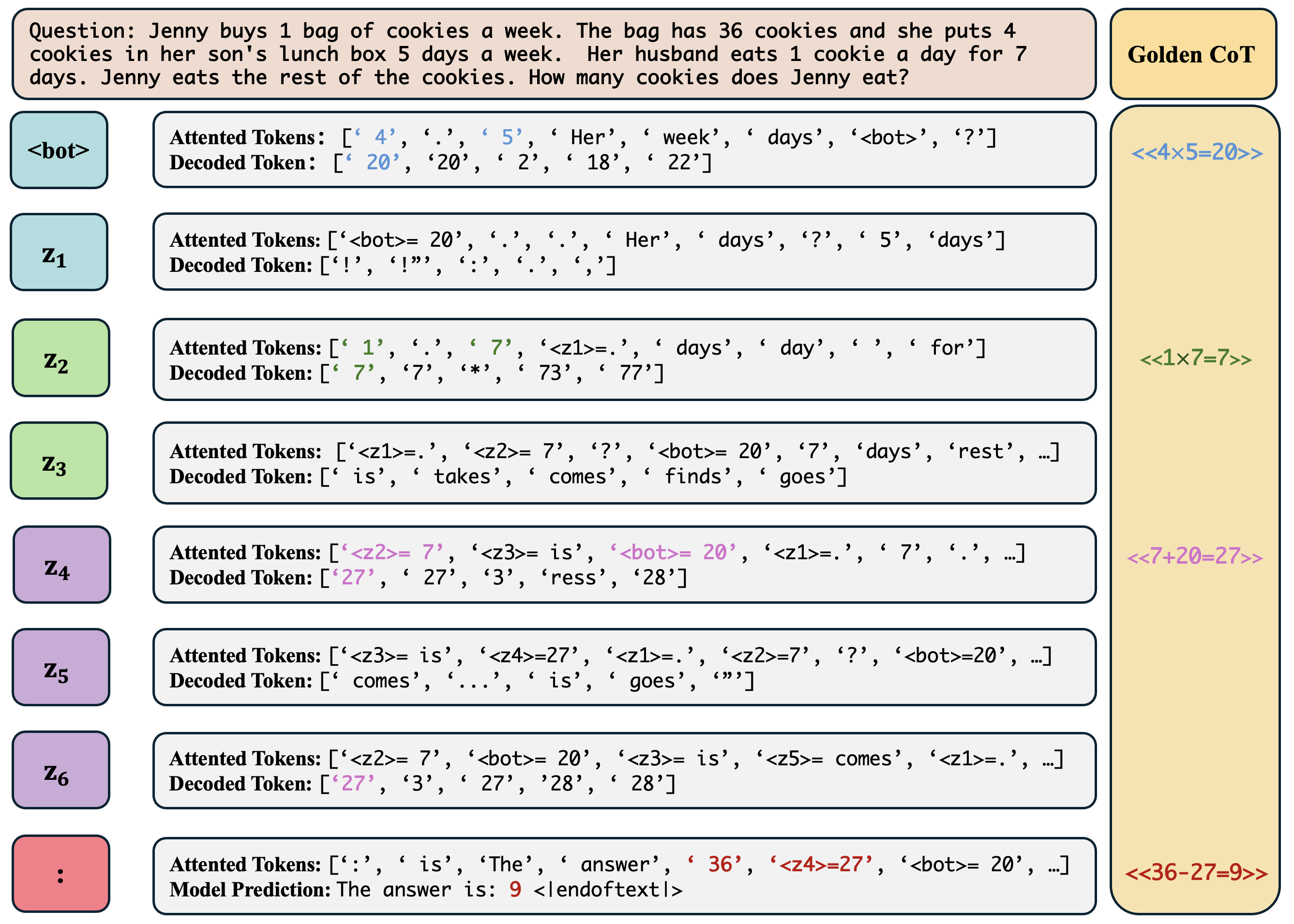}
  \centering
  \caption{%\footnotesize
  A case study illustrating CODI's interpretability by analyzing its attended tokens and decoded tokens of each of the six latent thought tokens, $z_1\cdots z_6$. \textbf{Attended tokens}: these represent the top-10 tokens that the continuous thought attends to when generating the next thought/token. Some attended tokens appear in the form of `\texttt{$z_i=x$}', indicating attention to the $i$-th continuous thought. Here $x$ represents the top-1 token that the latent thought maps to in vocabulary space. The model always attends to the first token in the sentence, so we remove that for better visualization.
  \textbf{Decoded tokens}: these are the top-5 words that the continuous thoughts are projected back to in vocabulary space by multiplying them with the vocabulary embeddings. }
  \label{fig:codi_interpretability}
  %\vspace{-10pt}
\end{figure*}

%\paragraph{Interpretability is hard in continuous space.}
%Interpreting CODI's continuous thoughts is inherently challenging because these representations lack explicit imitation targets. We interpret CODI's continuous thought by projecting its last hidden state into vocabulary space via the model's word embeddings -- treating it in the same way as a standard text token. 

\subsection{Interpretability Analysis}
Interpreting CODI's continuous thoughts is inherently challenging because these representations lack explicit imitation targets.
However, CODI exhibits an ability to produce observable intermediate results (Figure~\ref{fig:codi_interpretability}) within its continuous thoughts by projecting its last hidden state into vocabulary space via the model's word embeddings -- treating it in the same way as a standard text token. Additionally, the corresponding operands contributing to these intermediate results can often among the \textbf{top-ranked attended tokens} of the latent representation. For example, the second thought token, $z_2$, attends to both "1" and "7" to produce the decoded token "7". While the operator itself (e.g., $\times$) is not explicitly visible in the attention mechanism—since operators are in the context—it is reasonable to infer that the transformer layers \emph{implicitly} perform this operation. Another interesting observation is that each intermediate result is separated by a seemingly meaningless continuous token. We hypothesize that these tokens act as placeholders or transitional states during the computation of intermediate results. This aligns with the idea that the transformer may require multiple passes to complete the calculation for each intermediate step. More case studies are in the Appendix \ref{sec:interpretability_appendix}.

\begin{table}[h]
    \centering
    %\small
    %\resizebox{\columnwidth}{!}
    {
  \begin{tabular}{l|ccc}
    \toprule
    \textbf{Total Steps} & 1 & 2 & 3 \\
    \midrule
    \textbf{Accuracy} & 97.1\% & 83.9\% & 75.0\% \\
    \bottomrule
  \end{tabular}
  }
  \caption{%\footnotesize 
  CODI's top-5 intermediate results matching reference CoT across problems requiring different numbers of step.}
  \label{tab:interpretability}
  %\vspace{-10pt}
    
\end{table}
%\paragraph{Quantitative Evaluation.}
%\paragraph{Interpretability is a general pattern in CODI.}
Beyond the case study, we aim to establish that CODI’s interpretability is a general pattern by an accuracy metric. We extract all correctly predicted answers, decode the corresponding intermediate results, and compare them against the reference intermediate solutions.
Table~\ref{tab:interpretability} reveals that when there is only one intermediate result, CODI correctly matches the reference 97.1\% of the time. For CoT sequences with lengths up to 3, CODI consistently achieves over 75\% accuracy in decoding valid intermediate results.  These findings highlight CODI’s reliability in generating meaningful intermediate reasoning steps, demonstrating its potential to effectively handle reasoning tasks with interpretable intermediate outputs.

\section{Conclusion}
%\vspace{-5pt}
We introduced CODI, a novel paradigm for reasoning in continuous space. Our extensive experiments demonstrate CODI’s effectiveness as the new SOTA implicit CoT approach, while achieving a high compression ratio. Furthermore, CODI shows its robustness, generalisable to complex datasets, and interpretability. Future research should explore CODI’s application to more diverse and challenging tasks. 
%A promising direction is the integration of multimodality, leveraging continuous representations for seamless modality merging. 
We hope this work inspires further exploration into reasoning in representations more compact and robust than language, paving the way for more efficient and versatile reasoning paradigms.

\section{Limitations}
Implicit CoT methods inherently trade off interpretability compared to explicit CoT. While CODI provides a straightforward probing mechanism for inspecting continuous thoughts, it operates at the token level and faces limitations in reconstructing multi-token entities. For instance, a rare number like 35649 may span multiple tokens due to the tokenizer's behavior, but the current probing technique only decodes the first token, leaving the remaining components unobserved. More sophisticated probing techniques may be necessary to recover and visualize full semantic units.

Moreover, our approach focuses on knowledge transfer by probing the token (“:”) responsible for generating the first answer token. However, this choice may be suboptimal, as some answers begin with “-”, and removing such cases improves performance, suggesting that critical reasoning information might also reside in the token generating the second answer token. Additionally, probing the token that concludes the CoT reasoning—potentially summarizing the entire process—could offer alternative supervision signals. Furthermore, the current answer prompt, “The answer is:”, is an arbitrary design choice that may influence the effectiveness of knowledge transfer. Investigating these aspects further could enable CODI to extend its distillation framework to broader reasoning tasks.

Another limitation of the current continuous training approach is the absence of intermediate gradients until the end of the sequence. With six continuous thought tokens, the first token’s gradient is backpropagated from six or more steps away (specifically, from the token generating the final answer), which may introduce optimization challenges. This issue could become more pronounced when scaling to more complex problems requiring longer continuous reasoning chains.

%Furthermore, CODI’s current configuration is fully deterministic, whereas one advantage of CoT is its inherent stochasticity, which allows models to improve performance through self-consistency, tree search, and Monte Carlo Tree Search (MCTS). Introducing controlled randomness into CODI could restore this property. Potential solutions include integrating dropout layers, leveraging Variational Autoencoders (VAEs), or applying diffusion models atop the current projection layer to enable diverse and robust reasoning paths.

Finally, while we don’t have sufficient computation resources to scale the training of CODI on larger models, a concurrent paper \cite{geiping2025scalingtesttimecomputelatent} has demonstrated the feasibility of scaling a latent reasoning model to 3.5B parameters and 800 billion tokens with 4096 GPUs. The resulting model appears to be learning meta-strategies and abstractions for problem solving, as opposed to memorising as in existing LLMs trained on explicit CoT data. This is particularly encouraging, since not all reasoning steps can be easily verbalised (such as visual-spatial reasoning, emotional and social reasoning, and motor reasoning). While \citet{geiping2025scalingtesttimecomputelatent} focuses on pre-training, we proposed an efficient fine-tuning approach for equipping existing pre-trained LLMs with latent reasoning capabilities.
%Finally, our studies primarily focus on mathematical problems, as mathematical CoT training data are more abundant and mathematical problems inherently require complex reasoning. However, this focus may limit the generalizability of CODI to broader reasoning scenarios. Future work should explore its applicability to more diverse reasoning tasks beyond mathematical domains.

\section*{Acknowledgments}
This work was supported in part by the UK Engineering and Physical Sciences Research Council (EPSRC) (grant no. EP/V020579/1, EP/V020579/2, EP/Y003187/1, UKRI566, UKRI849). ZS is supported by a PhD studentship provided by the Chinese Scholarship Council.  The authors acknowledge the use of King's Computational Research, Engineering and Technology Environment (CREATE) at King's College London. We thank Lin Gui for his suggestions during both the submission and rebuttal stages of this paper.

\bibliography{acl_latex}

\clearpage
\appendix

\setcounter{table}{0}
\renewcommand{\thetable}{A\arabic{table}}
\setcounter{figure}{0}
\renewcommand{\thefigure}{A\arabic{figure}}

%\section{Appendix}
%\label{sec:appendix}

\section{Implementation Details}
\label{sec:hyperparams}
For all experiments (CoT-SFT, No-CoT-SFT, and CODI) on both GSM8K and Commonsense, we use the AdamW optimizer \cite{loshchilov2019decoupledweightdecayregularization} with a cosine scheduler (without cycles) and a linear warm-up over the first 3\% of steps. The effective batch size is 128. Both $\alpha$ and $\beta$ are set to 1 (Equation \ref{equ:total_loss}). We apply LoRA \cite{hu2021loralowrankadaptationlarge} finetuning with a rank of 128 and an alpha value of 32, using bfloat16 precision.

For GPT-2, we set the learning rate to 3e-3 and $\gamma$ to 1. Training runs for 40 epochs, taking approximately 36 hours on a single A100 (80GB).

For LLaMA-3.2-1b, we use a learning rate of 8e-4 and set $\gamma$ to 20, as we observe that its distillation loss has a much smaller magnitude. The model is trained for 10 epochs, requiring approximately 48 hours on a single A100 (80GB).

For iCoT training of GPT-2, we use a learning rate of 5e-5 and train for 100 epochs, removing 4 tokens per epoch for GSM8k-Aug-NL. For iCoT training of LLaMA-1b, we use a learning rate of 1e-5 and train for 50 epochs, removing 8 tokens per epoch for GSM8k-Aug and 16 tokens per epoch for GSM8k-Aug-NL. LoRA is not used during training.

For Coconut training of GPT-2, we use a learning rate of 1e-4 and train for 25 epochs without continuous tokens and 25 epochs with continuous tokens (50 epochs in total). For iCoT training of LLaMA-1b, we use a learning rate of 1e-5 and train 5 epochs for both stages (10 epochs in total). LoRA is not used during training.

\section{Proof: CoTs Contribute a Shift in Hidden Activation}
\label{appx:proof}
In this section, we provide a proof to demonstrate why Chain-of-Thought (CoT) contributes a shift in hidden activation. This proof is largely inspired by the work of \cite{li2024incontextlearningstatevector}, which analyzed In-Context Learning.

In a typical CoT training dataset, the input usually consists of four components: the question $Q$, the rationale $R$, the prompt for the answer $P$ (e.g., "The answer is:"), and the final answer $A$.

We analyze the attention activation of the last prompt token, $\textbf{q}$—in this case, ":"—at the $l$-th transformer layer. The output activation $\mathbf{a}^l$ from the attention heads of this token is given by:

\begin{equation}
    \mathbf{a}^l = W_V[Q; R; P]\text{softmax}(\frac{W_K[Q;R;P]^T\textbf{q}}{\sqrt{d}}) 
\end{equation}

where $W_K$ and $W_V$ are the model's key and value parameters, $[Q;R;P]$ represents the concatenation of the three inputs, and $\sqrt{d}$ is a scaling factor.

For simplicity of analysis, inspired by \cite{li2024incontextlearningstatevector}, we omit the softmax operation and the scaling factor, as these do not affect the core conclusion. With this simplification, the following derivation holds:

\begin{align*}
    \mathbf{a}^l &\approx W_V[Q; R; P] W_K[Q;R;P]^T\textbf{q} \\
    &= \Big(W_V Q(W_V Q)^T + W_V R(W_V R)^T \\
    &\quad \quad \quad \quad+ W_V P(W_V P)^T\Big)\textbf{q} \\
    &= \Big(W_V [Q;P](W_V [Q;P])^T \\
    &\quad \quad \quad \quad + W_V R(W_V R)^T\Big)\textbf{q} \\
    &= \Big(W_{\text{no-CoT}} + W_V R(W_K R)^T\Big)\textbf{q} \\
    &= \mathbf{a}^l_{\text{no-CoT}} + W_V R(W_K R)^T\textbf{q}
\end{align*}

Here, $W_{\text{no-CoT}}$ is defined as $W_V [Q; P](W_K [Q; P])^T$, accounting for the contribution of $Q$ and $P$ without the CoT rationale. Correspondingly, $\mathbf{a}^l_{\text{no-CoT}}$ represents the attention activation excluding CoT.

The additional term $W_V R (W_K R)^T\textbf{q}$ represents the contribution of the CoT rationale $R$ to the hidden activation. We can get the hidden activation by transforming the attention activation by a non-linear function $f$:

\begin{equation}
\mathbf{h}^l \approx \mathbf{h}^l_{\text{no-CoT}} + f\Big(W_V R(W_K R)^T\textbf{q}\Big)
\end{equation}

Thus, we conclude that the rationale $R$ in the CoT primarily contributes a shift in hidden activation values, emphasizing its role as an additive factor in the latent representation. This shift can be effectively captured and learned using a distance metric.

\section{Datasets}
\label{sec:data_appendix}
We provide examples and statistics of training datasets and evaluation benchmarks.

\subsection{Examples}

\begin{tcolorbox}[
    colback=white,            % Background color of the main box
    colframe=black!80,        % Border color
    arc=5mm,                  % Rounded corners
    coltitle=white,           % Text color for the title
    colbacktitle=black!60,    % Background color for the title
    fonttitle=\bfseries,      % Bold font for title
    title={GSM8k-Aug}             % Title text
]
\texttt{Question = "Out of 600 employees in a company, 30\% got promoted while 10\% received bonus. How many employees did not get either a promotion or a bonus?"} 

\texttt{CoT = "<<600*30/100=180>> <<600*10/100=60>> <<180+60=240>> <<600-240=360>>"}

\texttt{Answer = "360"}
\end{tcolorbox}

\begin{tcolorbox}[
    colback=white,            % Background color of the main box
    colframe=black!80,        % Border color
    arc=5mm,                  % Rounded corners
    coltitle=white,           % Text color for the title
    colbacktitle=black!60,    % Background color for the title
    fonttitle=\bfseries,      % Bold font for title
    title={GSM8k-Aug-NL}             % Title text
]
\texttt{Question = "Jen shared a pack of chocolates among her friends. She gave 20\% to Lucy, 30\% to Sarah and the remaining were shared equally among four others. If the pack contained 100 chocolates, how many chocolates were each of the four others getting?"} 

\texttt{CoT = "The total percentage given to Lucy and Sarah is 20\% + 30\% = 50\%. So, the remaining percentage that was shared among the others is 100\% - 50\% = 50\%. The total number of chocolates shared among the others is 100 * 50 / 100 = 50 chocolates. So, each of the four others received 50 / 4 = 12.5 chocolates."}

\texttt{Answer = "12.5"}
\end{tcolorbox}

\begin{tcolorbox}[
    colback=white,            % Background color of the main box
    colframe=black!80,        % Border color
    arc=5mm,                  % Rounded corners
    coltitle=white,           % Text color for the title
    colbacktitle=black!60,    % Background color for the title
    fonttitle=\bfseries,      % Bold font for title
    title={CommonsenseQA-CoT}             % Title text
]
\texttt{Question: "The sanctions against the school were a punishing blow, and they seemed to what the efforts the school had made to change?
Choices:
A: ignore
B: enforce
C: authoritarian
D: yell at
E: avoid"} 

\texttt{CoT = "The context of the sentence indicates that the sanctions are undermining or dismissing the efforts made by the school to change. The word "ignore" fits best here, as it conveys the sense of the sanctions not acknowledging the school's efforts."}

\texttt{Answer = "A"}
\end{tcolorbox}

\begin{tcolorbox}[
    colback=white,            % Background color of the main box
    colframe=black!80,        % Border color
    arc=5mm,                  % Rounded corners
    coltitle=white,           % Text color for the title
    colbacktitle=black!60,    % Background color for the title
    fonttitle=\bfseries,      % Bold font for title
    title={SVAMP}             % Title text
]
\texttt{Question = "There are 87 oranges and 290 bananas in Philip's collection. If the bananas are organized into 2 groups and oranges are organized into 93 groups. How big is each group of bananas?"} 
\texttt{Answer = "145"}
\end{tcolorbox}

\begin{tcolorbox}[
    colback=white,            % Background color of the main box
    colframe=black!80,        % Border color
    arc=5mm,                  % Rounded corners
    coltitle=white,           % Text color for the title
    colbacktitle=black!60,    % Background color for the title
    fonttitle=\bfseries,      % Bold font for title
    title={MultiArith}             % Title text
]
\texttt{Question = "There are 64 students trying out for the school's trivia teams. If 36 of them didn't get picked for the team and the rest were put into 4 groups, how many students would be in each group?"} 
\texttt{Answer = "7"}
\end{tcolorbox}

\begin{tcolorbox}[
    colback=white,            % Background color of the main box
    colframe=black!80,        % Border color
    arc=5mm,                  % Rounded corners
    coltitle=white,           % Text color for the title
    colbacktitle=black!60,    % Background color for the title
    fonttitle=\bfseries,      % Bold font for title
    title={GSM-Hard}             % Title text
]
\texttt{Question = "Janet’s ducks lay 16 eggs per day. She eats three for breakfast every morning and bakes muffins for her friends every day with 4933828. She sells the remainder at the farmers' market daily for \$2 per fresh duck egg. How much in dollars does she make every day at the farmers' market?"} 
\texttt{Answer = "-9867630.0"}
\end{tcolorbox}

\subsection{Statistics}
The statistics of training data are shown in Table \ref{tab:train_data_stats}, and the statistics of evaluation benchmarks are shown in Table \ref{tab:test_data_stats}.

\begin{table}[h]
    \centering
    \small
  \begin{tabular}{l|cc}
    \toprule
    \textbf{Training Dataset} & Num. Data & Avg. CoT Tokens \\
    \midrule
    GSM8k-Aug & 385,620  & 20.3\\
    GSM8k-Aug-NL & 384,625 & 49.0\\
    CommonsenseQA-CoT & 8,096 & 85.0 \\
    \bottomrule
  \end{tabular}
  \caption{Training data statistics.}
  \label{tab:train_data_stats}
\end{table}

\begin{table}[h]
    \centering
  \begin{tabular}{l|c}
    \toprule
    \textbf{Evaluation Benchmark} &  Data Size \\
    \midrule
    GSM8k & 1,319 \\
    SVAMP & 1,000 \\
    GSM-Hard & 1,319 \\
    MultiArith & 500 \\
    CommonsenseQA & 1,221 \\
    \bottomrule
  \end{tabular}
  \caption{Evaluation Benchmark statistics.}
  \label{tab:test_data_stats}
\end{table}

\section{CODI's Pattern Learning}
\label{sec:pattern}
\begin{table}[h]
\vspace{-5pt}
    \centering
    \resizebox{0.9\columnwidth}{!}{
  \begin{tabular}{l|ccccc}
    \toprule
    \textbf{GPT-2} & No-CoT-SFT & CODI & Coconut & Res &  Op-Res \\
    \midrule
    \textbf{Accuracy} & 19.1\% & 43.7\% & 34.1\% & 34.0\% & 35.7\%  \\
    %     \textbf{GPT-2} & No-CoT-SFT  & Res &  Op-Res \\
    % \midrule
    % \textbf{Accuracy} & 19.1\% & 34.0\% & 35.7\%  \\
    \bottomrule
  \end{tabular}
  }
    \caption{Comparison of GPT-2 finetuned on two datasets derived from CODI's decoded thoughts. \textbf{Res}: using intermediate results as CoT. \textbf{Op-Res}: using intermediate operators and results as CoT.}
  \label{tab:manual_ds}
  \vspace{-10pt}
\end{table}

Given that CODI’s continuous thoughts can often be decoded into intermediate results, it raises a question: is CODI effectively equivalent to a GPT-2 fine-tuned on a dataset containing CODI's decoded patterns? We created a dataset containing only intermediate results (e.g., “\texttt{CoT: 20, 7, 27. Result: 9}” translated from the case study in Figure \ref{fig:codi_interpretability}). Additionally, since some cases of CODI show decoded operators like `$\times$' and `$-$' interleaved with intermediate results, we also create a synthetic CoT dataset that includes both operators and results (e.g., ``\texttt{CoT: $\times$, 20, $\times$, 7, $+$, 27. Result: 9}'').
As shown in Table~\ref{tab:manual_ds}, while models trained on the two synthetic datasets outperform the No-\texttt{CoT-SFT} baseline, they perform much worse compared to CODI, though perform on par with Coconut. These result suggest that CODI learns richer information from the teacher task through distillation than pure imitation on language-level intermediate results alone,  highlighting the advantages of our training framework.

\section{Interpretability Case Studies}
\label{sec:interpretability_appendix}
\begin{figure*}[t]
\small
  \includegraphics[width=0.9\textwidth, keepaspectratio]{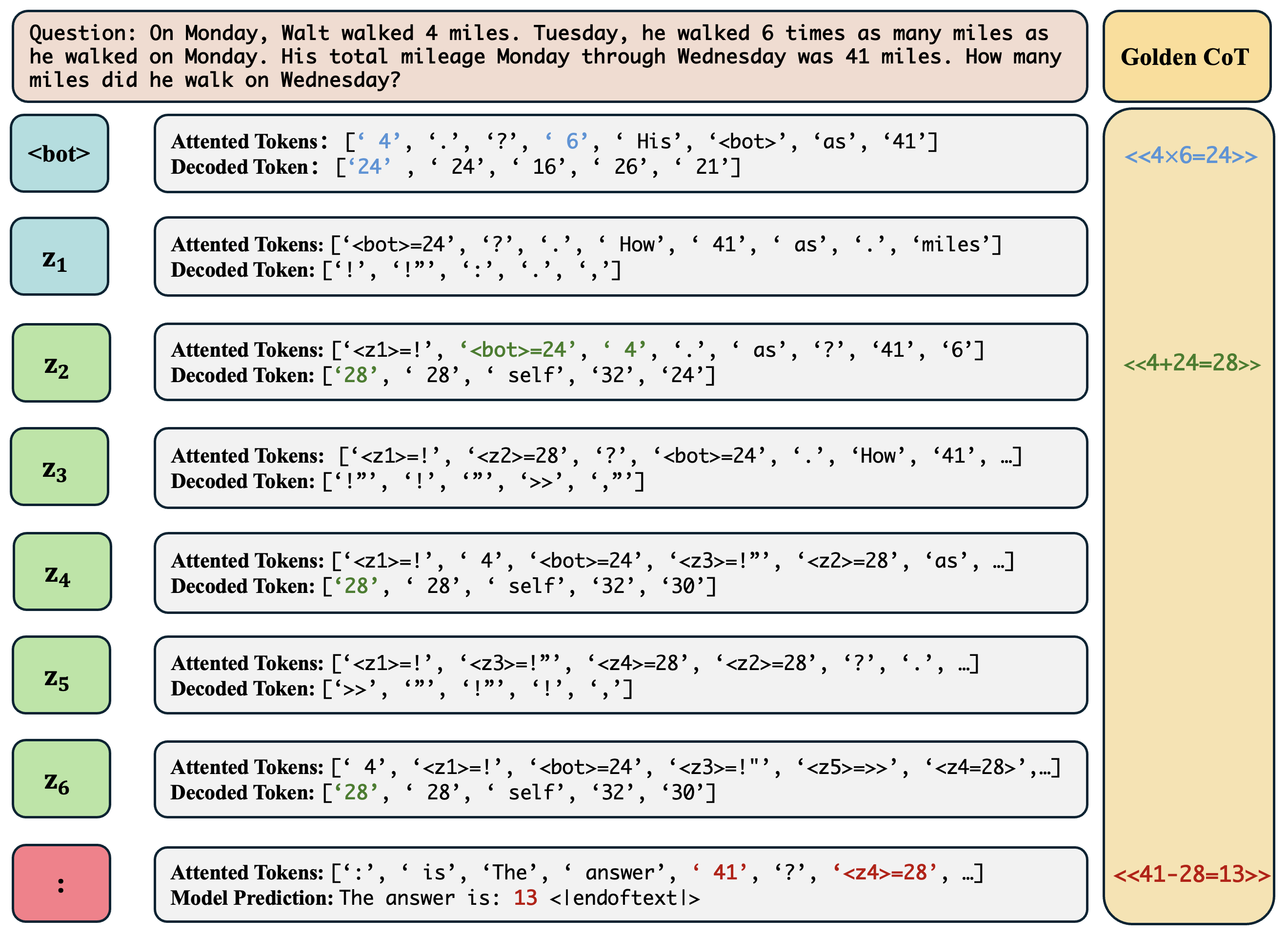}
  \centering
  \caption{CODI's interpretability on problems involving two steps.}
  \label{fig:codi_interpretability_2steps}
  \vspace{-5pt}
\end{figure*}

\begin{figure*}[t]
\small
  \includegraphics[width=0.9\textwidth, keepaspectratio]{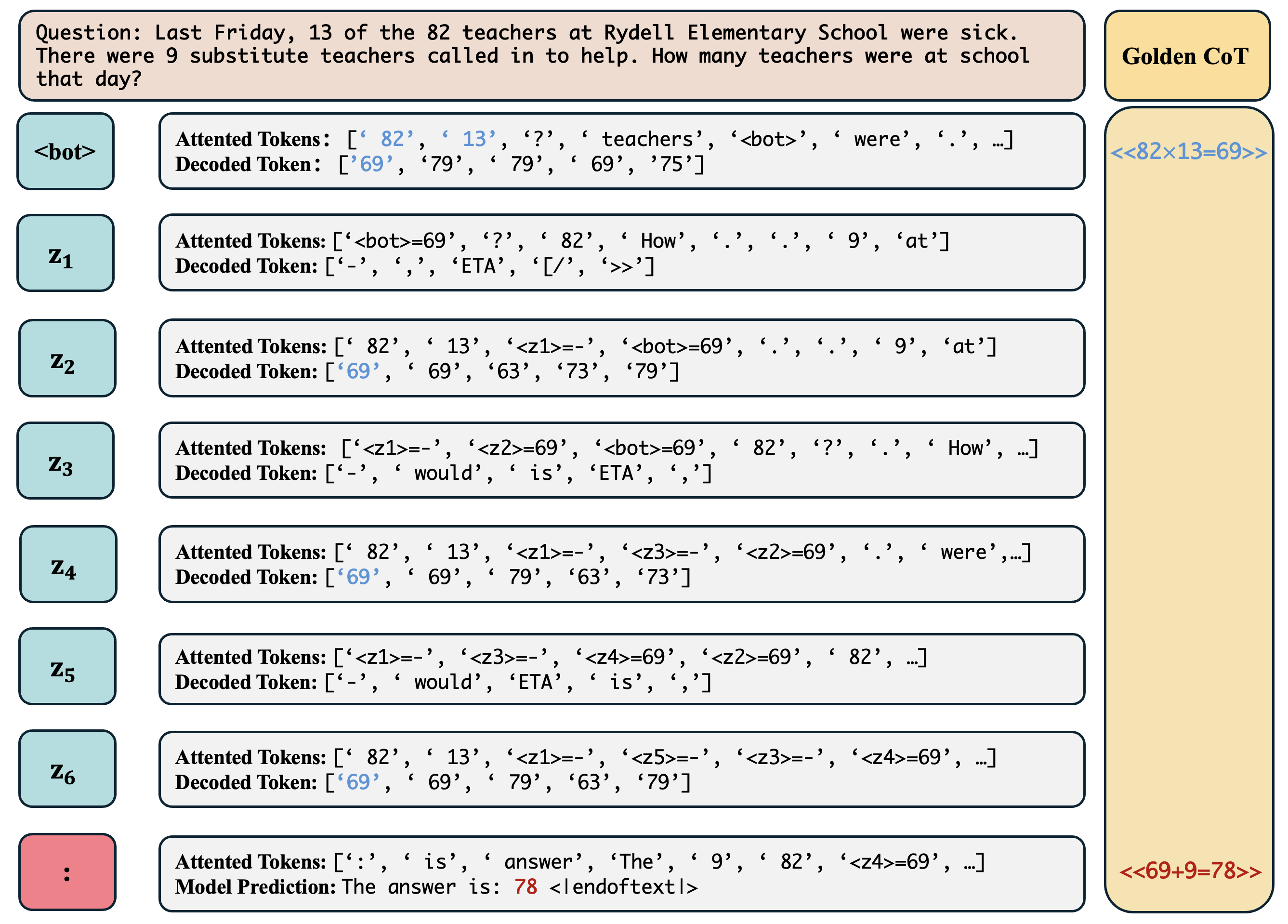}
  \centering
  \caption{CODI's interpretability on problems involving one step.}
  \label{fig:codi_interpretability_1step}
  \vspace{-5pt}
\end{figure*}

More case studies on the interpretability of CODI are provided in Figure \ref{fig:codi_interpretability_2steps} and Figure \ref{fig:codi_interpretability_1step}

\section{Ablations on the Hyperparameter}
\label{sec:ablation_hyperparameters}
The default settings for $\alpha$, $\beta$, and $\gamma$ from Equation \ref{equ:total_loss} are 1, and we fix $\alpha=1$ for the ablations below.

$\beta$ determines the weight of the distillation loss. We find that $\beta=1$ works well for GPT-2. However, for LLaMA models, the magnitude of the distillation loss is about 10 times smaller than in GPT-2, prompting us to test larger values of $\beta$. From Table \ref{tab:ablation_beta}, increasing $\beta$ from 1 to 5 leads to a substantial accuracy improvement. Beyond $\beta$ = 5, performance plateaus, remaining stable as $\beta$ increases up to 30. Therefore, our choice of $\beta$ for LLaMA-1b is aligned with the relative scale of the distillation loss. Based on this ablation, we select $\beta$ = 20 as the default value for LLaMA-1b.

$\gamma$ determines the relative weight between the explicit CoT reasoning objective (teacher task) and the implicit CoT objective (student task) during training. Table \ref{tab:ablation_gamma} shows that a higher $\gamma$ accelerates convergence but leads to lower final performance. This likely occurs because a larger $\gamma$ encourages the model to learn more from natural language CoT reasoning (the teacher task), which serves as the main source for developing its reasoning ability and thus improves early training performance. However, since the model is ultimately evaluated on implicit CoT (the student task), which receives less emphasis during training when $\gamma$ is large, its performance on the target objective declines.

\begin{table}[h]
    \centering
    \small
  \begin{tabular}{l|ccccc}
    \toprule
    \textbf{$\beta$} &  1 & 5 & 10 & 20 & 30 \\
    \midrule
    \textbf{Accuracy} & 46.5\% & 50.2\% & 49.1\% & 51.9\% & 51.4\% \\
    \bottomrule
  \end{tabular}
  \caption{Ablation study on $\beta$ on LLaMA-1b and GSM8k-Aug.}
  \label{tab:ablation_beta}
\end{table}

\begin{table}[h]
    \centering
  \begin{tabular}{l|cc}
    \toprule
    \textbf{$\gamma$} &  20 epochs & 40 epochs \\
    \midrule
    0.5 & 36.3\% & 38.2\% \\
    1 & 38.4\% & 43.7\% \\
    2 & 41.6\% & 41.9\% \\
    3 & 40.8\% & - \\
    \bottomrule
  \end{tabular}
  \caption{Ablation study on $\gamma$ on GPT-2 and GSM8k-Aug. Results report accuracy (\%) after training for different numbers of epochs.}
  \label{tab:ablation_gamma}
\end{table}

\section{Ablations on the Choice of the Distillation Token.}
\label{sec:ablation_distill_token}

\begin{table*}[h]
    \small
    \centering
  \begin{tabular}{l|c|c|c|c}
    \toprule
    \textbf{ ID} & \textbf{Prompt Design} &  \textbf{Distillation Token}  & \textbf{Accuracy} & \textbf{Within ±2×std of baseline?} \\
    \midrule
    1 & The answer is: (baseline) & : & 39.0\% & -\\
    2 & Answer: & : & 38.4\% & Yes\\
    3 & Therefore, based on all previous calculations, && \\ & we conclude that the final answer is: & : &40.2\% & Yes \\
    4&The answer is & is & 38.1\% & Yes\\
    5&We give the answer as & as &40.1\% & Yes \\
    6&We find the answer to be & be &39.0\% & Yes \\
    7&The answer is boxed\{ & \{ &38.4\% & Yes \\

    \bottomrule
  \end{tabular}
  \caption{Robustness test on the answer prompt of CODI trained on GSM8k-Aug with 20 epochs.}
  \label{tab:ablations_answer_token}
\end{table*}

We have conducted ablation studies to evaluate CODI’s robustness to various distillation tokens and answer prompts. As shown in Table \ref{tab:ablations_answer_token}, we tested a diverse set of prompts: prompts 2–3 vary the language, while prompts 4–7 focus on different distillation tokens (the last token of the prompt). To determine whether the accuracy differences are statistically significant, we follow an informal t-test approach, considering results to be significant if they fall outside the interval of ±2×std (1.8) from the baseline mean (39\%), which are obtained by 5 independent runs. Our findings indicate that none of the alternative prompt designs show a statistically significant difference from the baseline, suggesting that CODI is robust to variations in both distillation tokens and answer prompt styles.

\section{CODI Code}
The example Python code of CODI is illustrated in Figure \ref{fig:code}.

\begin{figure*}[t]
\centering
\begin{lstlisting}
class ContinuousCoTviaKnowledgeDistillation:
	def __init__(self,):
		self.num_latent = 6
		self.alpha, self.beta, self.gamma = 1, 1, 1
        self.llm = get_gpt2_model()
		self.prj = nn.Sequential(
			nn.Linear(hidden_dim, hidden_dim),
			nn.GELU(),
            		nn.Linear(hidden_dim, hidden_dim), 
            		nn.LayerNorm(hidden_dim),
		)
	
	def forward(x, y, x_cot_y):
		# teacher learning
		y_teacher = self.llm(x_cot_y)
		teacher_ce_loss = cross_entropy(y_teacher, x_cot_y) # loss1
		
		# student learning
		latent = self.llm(torch.cat([x, bot_token], dim=1))[:, -1]
		latent = self.prj(latent)
		past_key_values = latent.past_key_values
		
		# continuous CoT reasoning
		for i in range(self.num_latent):
			latent = self.llm(latent, past_key_values)
			latent = self.prj(latent)
			past_key_values = latent.past_key_values
		
		y_student = self.llm(torch.cat([eot_token, y], dim=1), past_key_values)
		student_ce_loss = cross_entropy(y_student, y) # loss2
		
		# knowledge distillation
		knowledge_distillation_loss = smooth_l1_loss(
			y_teacher.hidden_states[:, teacher_exact_answer_token_position-1],
			y_student.hidden_states[:, student_exact_answer_token_position-1]
		) # loss3
		# normalisation
		knowledge_distillation_loss /= y_teacher.hidden_states[:, teacher_exact_answer_token_position-1].std()
		
		return self.alpha*student_ce_loss teacher_ce_loss + self.beta*knowledge_distillation_loss + self.gamma*teacher_ce_loss
        
\end{lstlisting}
\caption{Example Python code illustrating the ContinuousCoTviaKnowledgeDistillation class.}
\label{fig:code}
\end{figure*}

\end{document}